\documentclass[10pt,twocolumn,letterpaper]{article}

\usepackage[pagenumbers]{cvpr} 

\definecolor{cvprblue}{rgb}{0.21,0.49,0.74}
\usepackage[pagebackref,breaklinks,colorlinks,allcolors=cvprblue]{hyperref}

\usepackage[accsupp]{axessibility}  

\usepackage{booktabs} 
\usepackage{graphicx} 
\usepackage{colortbl} 
\usepackage[table]{xcolor} 
\usepackage{amsmath}  
\usepackage{multirow} 

\definecolor{best}{RGB}{255, 160, 160}  
\definecolor{second}{RGB}{255, 224, 192} 
\definecolor{third}{RGB}{255, 255, 204} 
\newcommand{\up}{\ensuremath{\uparrow}}
\newcommand{\down}{\ensuremath{\downarrow}}

\newcommand{\lyj}[1]{\textcolor{red}{lyj:#1}}
\newcommand{\chong}[1]{\textcolor{orange}{Chong: #1}}

\renewcommand{\lyj}[1]{}
\renewcommand{\chong}[1]{}

\usepackage{float} 

\def\paperID{*****} 
\def\confName{CVPR}
\def\confYear{2026}

\title{D-Prism: Differentiable Primitives for Structured Dynamic Modeling}

\author{
Xingyuan Yu\textsuperscript{1} \quad 
Yijin Li\textsuperscript{1} \quad 
Chong Zeng\textsuperscript{2} \quad
Yuhang Ming\textsuperscript{3} \quad
Hujun Bao\textsuperscript{1} \quad 
Guofeng Zhang\textsuperscript{1\dag}\\
\hspace{-2.5mm}\textsuperscript{1}State Key Lab of CAD\&CG, Zhejiang University ~
\textsuperscript{2}Stanford University ~
\textsuperscript{3}Hangzhou Dianzi University\\
{\tt\small rickyyxy@zju.edu.cn, eugenelyj@foxmail.com, chongzeng@siggraph.org,}\\
{\tt\small yuhang.ming@hdu.edu.cn, bao@cad.zju.edu.cn, zhangguofeng@zju.edu.cn}
}

\begin{document}
\twocolumn[{
\maketitle
\vspace{-1cm}
\begin{center}
    \includegraphics[width=\linewidth]{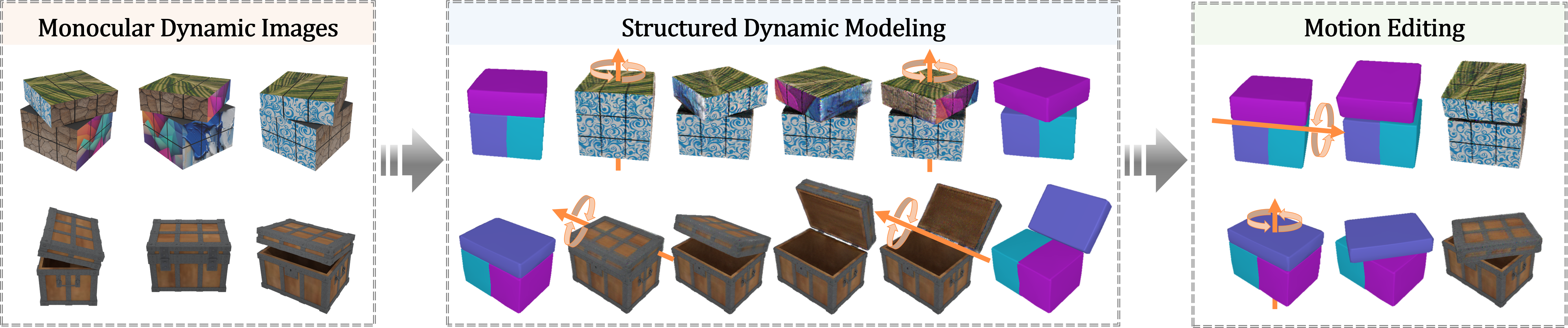}
\end{center}
\vspace{-0.5cm}
\captionsetup{type=figure}
\captionof{figure}{%
    \textbf{D-Prism} is a novel framework based on structured primitives for dynamic geometry reconstruction using monocular inputs. It demonstrates a superior capability for modeling dynamic structured objects, providing accurate part-based geometry and motion reconstruction, alongside high-quality appearance. The resulting reconstruction enables applications like motion editing, such as swapping motion patterns between different objects.\lyj{Comments: Can we provide some better application examples? It looks like a simple reverse play.}\chong{Monocular Images $\rightarrow$ Monocular Dynamic Images, emphasize the input contains dynamic. Agree with lyj that we need to show more complex applications.}
}\label{fig:teaser}
\vspace{0.3cm}
}]


\begin{abstract}
Capturing both geometry and rigid motion\chong{does our method reconstruct those "constraints" beside rigidity? Like knob collision etc. if not, might be dangerous to mention this, or you want to scope down the term "constrained motion"} for structured dynamic objects, like multi-part assemblies or jointed mechanisms, remains a key challenge. 
Existing dynamic methods, such as deformable\chong{deformable?} meshes or 3DGS, rely on unstructured representations and fail to jointly model suitable geometry and articulated motion. Primitive-based methods excel at structured static scenes, but their dynamic potential is still unexplored. We propose D-Prism, the first framework to achieve high-fidelity structured dynamic modeling by extending differentiable primitives to the dynamic domain.
Specifically, we bind 3DGS to primitive surfaces, leveraging their respective strengths in appearance and geometry. We introduce a deformation network to control primitive motion, ensuring it accurately matches the object's movement. 
Furthermore, we design a novel adaptive control strategy to dynamically adjust primitive counts, better matching objects' true spatial footprint.
Experiments confirm that our method excels at structured dynamic modeling, providing both structured geometry and precise motion tracking. Project page: \url{https://zju3dv.github.io/d-prism/}.

\end{abstract}    
\vspace{-0.4cm}
\section{Introduction}
\label{sec:intro}

Creating structured representations for objects \cite{marr2010vision} is a crucial task, as many real-world objects are inherently structured as multi-part assemblies or articulated mechanisms, from industrial robotics to everyday tools. A common requirement is to dynamically model these objects from visual input, which requires recovering both their part geometry and rigid motion.\lyj{I would prefer to say ``a common scene or a common requirement'' instead of ``a key challenge'' here. Besides, what do you mean ``constrained motion''.} This provides a fundamental prior for high-impact applications, including scene editing, scene understanding, and embodied AI.

While many methods have explored dynamic scene reconstruction, the majority \cite{pumarola2021d,cao2023hexplane,wu20244d} concentrate on the Novel View Synthesis (NVS) task and cannot provide an explicit geometric representation. Other methods \cite{liu2024dynamic,johnson2022ub4d} explore geometry by representing the entire dynamic object as a single, continuous dynamic mesh. However, this monolithic representation is fundamentally flawed, as it lacks an intrinsic, part-based decomposition and thus fails to model structured objects. Furthermore, it struggles with geometry degradation during part interactions, such as proximity or separation. For example, a rotating Rubik's Cube face as shown in Fig. \ref{fig:teaser}\chong{as shown in fig xxx} should be modeled as two parts, which is a circumstance that current methods cannot handle.

To address these problems, we propose D-Prism, a novel framework for modeling dynamic objects using a set of primitives. While primitive-based methods have shown success in static scene reconstruction \cite{monnier2023differentiable,jiang2024gaussianblock,gao2025self}, yielding a compact, actionable, and interpretable representation, we innovatively introduce this structured approach to dynamic situations. We use its inherent part-based nature to solve cases previous methods could not handle, such as the motions of dynamic structured objects, including contact, separation, and hinged movements. By accurately modeling these motions, our approach reveals and restores the true underlying structure of objects, including their structure at rest and at any moment during movement.
Furthermore, we bind 3D Gaussians \cite{kerbl20233d} to the primitives, allowing us to simultaneously achieve a structured geometric representation and high-quality appearance reconstruction.

Nonetheless, making primitives move as coherent parts of a dynamic object introduces a complex optimization problem. 
Aside from modeling motion for the primitives, a more fundamental difficulty is that the instability of primitive optimization under sparse viewpoints is severely compounded in monocular dynamic scenes, where observations of moving parts become even sparser.
This problem is also exacerbated by the overly simplistic adaptive strategies from static methods which are unable to balance redundancy and completeness for moving objects, often leading to a degradation in modeling quality for the primitives or even to training failure.
To overcome these limitations, our framework introduces deformation networks \cite{pumarola2021d} to model primitive motion and a dynamic adaptive control strategy to manage the primitive set.\chong{I don't see the connection between deformation network and the camera motion, for mono input recon, camera motion = object motion. Also I don't see the relation to initialization.} 
The deformation network enables more stable learning of primitive motion, mapping primitives between the canonical space and any observation space. This differentiable approach allows for the joint optimization of primitive motion and other primitive parameters.
In parallel, unlike the simple and limited strategies of static methods \cite{monnier2023differentiable,gao2025self}, the dynamic adaptive control strategy manages the primitive set by autonomously adding, merging, or removing primitives.
This policy significantly enhances our framework's robustness, enabling it to represent diverse object types and reconstruct the various motion patterns of these objects.
We also ensure that primitives updated by this process are consistent in both the canonical and observation spaces. 

Finally, we create the Dynamic Primitive Dataset using assets from PartNet-Mobility Dataset \cite{Xiang_2020_SAPIEN} to specifically evaluate our method's effectiveness on structured objects. Concurrently, we introduce structured motion tracking accuracy, a new metric to better assess true motion recovery in reconstruction. 
We also test our method in more general scenes, such as cases in D-NeRF Dataset \cite{pumarola2021d} and real-world scenes.\chong{and real-world captured videos?} The overall experimental results demonstrate that our approach can better reconstruct object motion while providing an accurate, structured geometric representation. 
We believe this is the first framework to reconstruct structured geometry while maintaining inter-frame consistency from dynamic monocular observations.
In summary, our contributions are as follows: 
\begin{itemize} 
\item A novel framework using structured primitives for dynamic geometry reconstruction, which provides part-based geometric results and a superior capability for modeling dynamic structured objects.
\item Multiple training strategies for dynamic primitive modeling that capture the object part motion while maintaining geometric integrity. 
\item A new dataset of structured objects and a specialized metric that together show improved reconstruction of dynamic structured objects over previous methods.
\end{itemize}

\section{Related Work}
\label{sec:related_work}

\subsection{Primitive-based Representation}
Decomposing scenes into a set of geometric primitives is a fundamental approach for enabling high-level tasks such as semantic understanding and scene editing. The origins of this concept can be traced to seminal work such as Blocks World \cite{DBLP:books/garland/Roberts63} in the 1960s and Generalized Cylinders \cite{binford1975visual} in the 1970s. This field has since evolved to include a variety of geometric primitives, such as cuboids \cite{ramamonjisoa2022monteboxfinder,tulsiani2017learning}, superquadrics \cite{liu2022robust,DBLP:conf/cvpr/PaschalidouUG19,paschalidou2020learning,wu2022primitive}, and convex shapes \cite{chen2020bsp,deng2020cvxnet}. Notable examples include MonteBoxFinder \cite{ramamonjisoa2022monteboxfinder}, which integrates clustering, cuboid fitting, and Monte Carlo Tree Search for scene parsing, and EMS \cite{liu2022robust}, which uses a probabilistic approach for robust superquadric recovery. The pursuit of more detailed representations has also led to research into flexible primitive deformations \cite{gao2019sdm,hui2022neural,liu2023deformer,paschalidou2021neural,shuai2023dpf,fedele2025superdec}.
A major trend in recent years is the development of methods that learn structure-aware 3D representations directly from 2D images \cite{monnier2023differentiable, zhou2025monomobility}. This line of work includes methods such as ISCO \cite{alaniz2023iterative}, PartNeRF \cite{tertikas2023generating} and DBW \cite{monnier2023differentiable}. They have recently been extended by approaches like PartGS \cite{gao2025self} and GaussianBlock \cite{jiang2024gaussianblock} which leverage 3D Gaussians for primitive appearance.

However, a focus solely on static scenes is insufficient, as an object's true structure is often revealed only through its motion. For example, a closed chest appears as a single cuboid until it opens. Our work addresses this critical gap by being the first to apply the structured representation to the dynamic reconstruction pipeline and thus broadens the application domain of primitives.

\begin{figure*}[!t]
  \centering
   \includegraphics[width=1\linewidth]{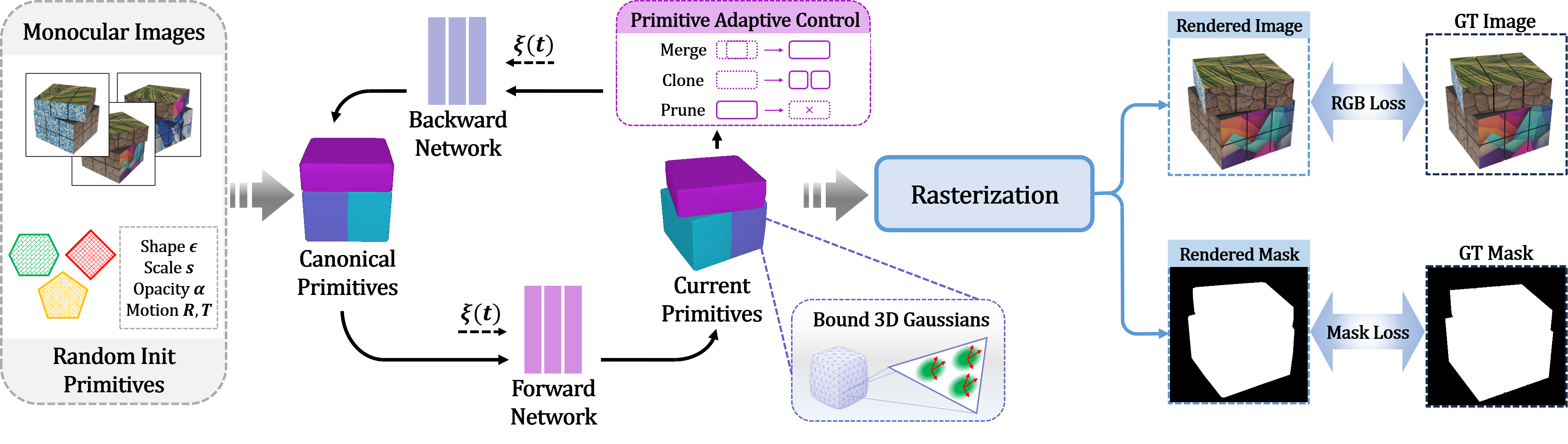}
   \caption{\textbf{Overview of D-Prism.} Given calibrated monocular images and masks, our method learns the structured dynamic geometry and appearance for the sequence. Deformation networks model the object's underlying motion and drive the primitives, while the primitive adaptive control strategy manages their count and distribution to enhance our framework's representational ability.\chong{rendering mask $\rightarrow$ rendered mask}\chong{if the primitive is superquadric/icosphere, then the bound 3D gaussians part should also use that as the primitive's shape} \chong{Forward Network $\rightarrow$ Forward Deform Network}}
   \label{fig:framework}
   \vspace{-0.5cm}
\end{figure*}

\subsection{Dynamic Geometry Reconstruction}
Dynamic reconstruction is a key problem in computer vision, crucial for applications like virtual reality, digital twins, and robotic interaction. In the past, many methods have tackled this problem at the level of Novel View Synthesis (NVS) \cite{wu2024recent}, including NeRF-based approaches \cite{park2021nerfies,park2021hypernerf,du2021neural,gao2021dynamic,li2021neural,xian2021space,pumarola2021d,jing2025frnerf}, 4D volumetric methods \cite{cao2023hexplane,fridovich2023k,shao2023tensor4d,wei2025neumanifold,fang2022fast}, and recent deformed 3D Gaussian techniques \cite{yang2024deformable,wu20244d,luiten2024dynamic,li2024spacetime,yang2023real,wang2025shape}. However, NVS alone is insufficient for many applications that require a tangible geometric entity.

Reconstructing this explicit dynamic geometry is therefore a critical but distinct challenge. Compared to static mesh recovery \cite{hanocka2020point2mesh,xue2023nsf,pan2019deep,wang2018pixel2mesh,kanazawa2018learning,li2020online}, dynamic reconstruction must also address complex issues such as handling topological shifts as the object deforms. Recent works have utilized differentiable implicit representations \cite{wang2021neus,liao2018deep,chen2022neural,remelli2020meshsdf,shen2021deep,shen2023flexible}, which have enabled attempts to learn deformable geometry from monocular video \cite{liu2024dynamic,johnson2022ub4d,yang2021lasr,yang2023ppr,wu2023dove,tulsiani2020implicit}.

However, current monocular methods face a trade-off: dynamic meshes often sacrifice the ability to maintain consistent vertex correspondences to achieve high per-frame accuracy. Furthermore, existing approaches struggle with structured objects \cite{marr2010vision}, as a monolithic mesh cannot handle the complex kinematics of structured parts, such as surface contact and separation. Therefore, we are the first to introduce a structured representation into the dynamic geometry reconstruction, aiming to solve these specific problems.
\section{Method}

Given a set of monocular images $I_{1:N}$ of a dynamic object, with corresponding timestamps $t_{1:N}$, camera parameters $c_{1:N}$, and object masks $M_{1:N}$, we establish a set $\mathcal{S}=\{ P_1, \dots, P_K \}$ of $K$ primitives. We maintain the motion of these primitives across all timestamps using a primitive deformation network. Furthermore, each primitive has bound 3D Gaussians, which serve as the appearance of the primitive in place of traditional textures. 
This section details our framework, illustrated in Fig. \ref{fig:framework}. Sec. \ref{sec:deform_pri} describes the parameterization of our structured representation and the deformation network controlling primitive motion. Sec. \ref{sec:optim} details the optimization process, including the adaptive control strategy for managing primitive count and distribution. Finally, Sec. \ref{sec:refine} describes the refinement stage, where we further optimize parameters for high-quality appearance, more accurate shape and motion reconstruction.

\subsection{Deformable Structured Representation}
\label{sec:deform_pri}
\subsubsection{Differentiable Primitive Parameters}
We represent the object using a set of primitives $\mathcal{S}$. Each primitive $P_i$ is defined as a superquadric mesh, which offers strong expressiveness with only a few continuous parameters \cite{barr1981superquadrics,paschalidou2019superquadrics}, thereby ensuring the entire representation is differentiable.
Specifically, the base configuration of a primitive is a unit icosphere. A mapping function is used to transform the vertices of this icosphere to our desired positions, thereby defining the final primitive shape. If the vertices on the icosphere are defined by spherical coordinates $\eta \in [-\pi/2, \pi/2]$ and $\omega \in [-\pi, \pi]$, then this mapping function $\mathcal{F}$ can be expressed as the following equation \cite{barr1981superquadrics}:
\begin{equation}
    \mathcal{F}(\eta, \omega) = 
    \begin{bmatrix}
    s_1 \cos^{\epsilon_1} \eta \cos^{\epsilon_2} \omega \\
    s_2 \sin^{\epsilon_1} \eta \\
    s_3 \cos^{\epsilon_1} \eta \sin^{\epsilon_2} \omega
    \end{bmatrix},
\end{equation}
where $\epsilon_1$ and $\epsilon_2$ are two shape parameters, while $s_1, s_2, s_3$ are three scaling parameters. In addition, each primitive $P_i$ has motion parameters: a rotation $R_i \in \mathbb{R}^6$ and a translation $T_i \in \mathbb{R}^3$. It also has an opacity parameter $\alpha_i$. Here, $R_i$ employs the 6D rotation parametrization \cite{zhou2019continuity}, which is convertible between the 6D vector and its corresponding rotation matrix. These motion parameters map $P_i$ from its local space to the world space via the transformation: $x_{\text{world}} = \text{rot}(R_i)x + T_i$. All these parameters are learnable.\chong{should draw these properties in the illustration in fig2}

\subsubsection{Bound 3D Gaussians Parameters}
To optimize the primitive's appearance, we bind 3D Gaussians to its surface. At initialization, we randomly distribute Gaussian centers across the surface, defined by barycentric coordinates that temporarily remain fixed during the main training.\chong{temporarily fixed during the first stage, xxx} Similar to \cite{gao2025self,waczynska2024games}, each Gaussian’s rotation matrix $R_v$ and scaling $S_v$ are computed from vertex positions. Given the limited vertices on the icosphere, we choose three temporarily fixed Gaussian centers nearest to the current one as $v_1, v_2, v_3 \in \mathbb{R}^3$ rather than primitive vertices,\chong{does this mean the gaussian's property relies on other gaussian?} which enhances precision. The orthonormal vectors are then constructed: $r_1$ aligns with the face normal, $r_2$ is derived from the vector pointing from the centroid ($m = \text{mean}(v_1, v_2, v_3)$) to $v_1$, and $r_3$ is determined by orthogonalizing \cite{bjorck1994numerics} the vector from $m$ to $v_2$ with respect to both $r_1$ and $r_2$:
\begin{equation}
    R_v = [r_1,r_2,r_3], \quad r_3 = \frac{\text{ort}(v_2 - m; r_1, r_2)}{||\text{ort}(v_2 - m; r_1, r_2)||}.
\end{equation}
The scaling $S_v = \text{diag}(s_1, s_2, s_3)$ is determined by local geometry. The in-plane scales $s_2, s_3$ are based on the maximum distance $d_{\text{max}}$ from the center $g_c$ to its neighbors $v_1, v_2, v_3$, ensuring coverage of the local triangular patch. The normal-direction scale $s_1$ is set to a small constant $\tau_s = 1e^{-8}$ to represent a thin surface. The scaling $S_v$ is computed as:
\begin{equation}
    S_v = \text{diag}(\tau_s, d_{\text{max}}, d_{\text{max}}),
    \quad d_{\text{max}} = \max_{k \in \{1,2,3\}} ||v_k - g_c||.
\end{equation}
Additionally, the Gaussian's opacity is determined by its host primitive. These parameters are not independently optimized during the main training, only the SH coefficients that represent appearance are optimized separately.

\subsubsection{Deformation for Primitives}
Following previous work on dynamic reconstruction \cite{pumarola2021d,yang2023movingparts,li2024spacetime}, we also define a deformation network to transform the primitive $P(T;R,\epsilon,s)$ from one space to another:
\begin{equation}
    \mathcal{D}(\xi(T), \xi(t)) = (\Delta T, \Delta R).
\end{equation}
\chong{is $\Delta R$ also using the 6D parameterization as the base primitive?}Here $\Delta R \in \mathbb{R}^6, \Delta T \in \mathbb{R}^3$. $t$ denotes the timestamp and $\xi$ denotes the positional encoding \cite{tancik2020fourier} that projects $T$ and $t$ into a higher-dimensional Fourier space.\chong{add detail: how many frequencies are used in the encoding? and what freqs} Since each primitive corresponds to a fixed object part, we assume its shape and scale remain constant over time. Therefore, the deformation only updates the motion parameters, and the deformed primitive is represented as: $P(T+\Delta T;R+\Delta R,\epsilon,s)$. We also introduce an inverse network that maintains an identical input/output format and structure.

\begin{figure*}[!t]
  \centering
   \includegraphics[width=1\linewidth]{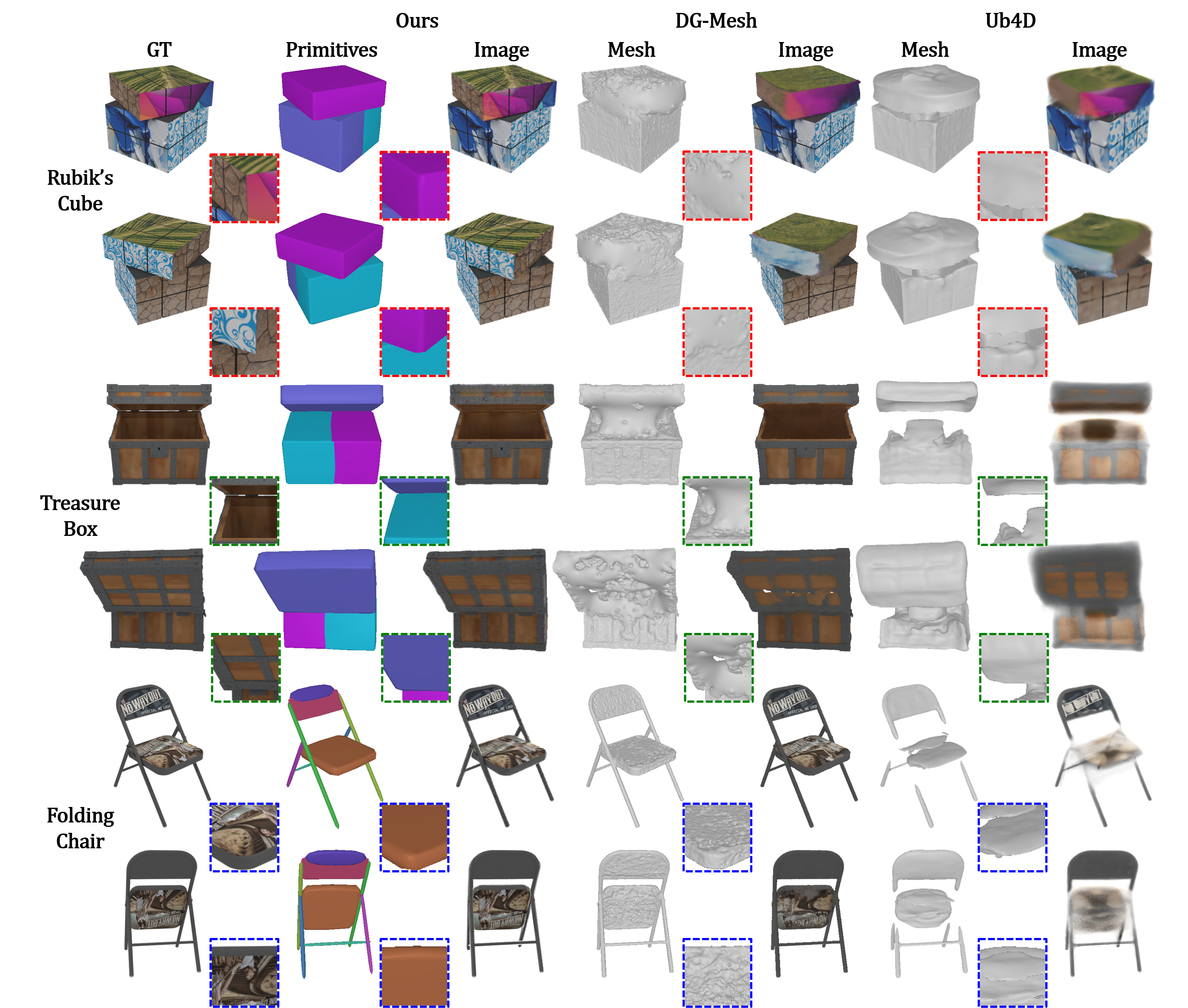}
   \caption{\textbf{Visual Comparison of Structured Dynamic Modeling.} We visualize the cases from Dynamic Primitive Dataset. We show both geometry reconstruction results and rendering images. Previous methods exhibit severe errors in geometric structure. In contrast, our method perfectly restores the dynamic object's structure and motion process.}
   \label{fig:mesh_vis}
\end{figure*}

\begin{table*}[!t] 
  \centering
  \tiny
  \caption{\textbf{Structured Motion Tracking Results on Dynamic Primitive Dataset.} We track the geometry structure in each frame. We then average the results across all frames. We use \textcolor{best}{\rule{1em}{1.5ex}}, \textcolor{second}{\rule{1em}{1.5ex}}, and \textcolor{third}{\rule{1em}{1.5ex}} to indicate the best, the second best and the third results. The results show our method excels at modeling structured dynamic objects. Our suitable structural decomposition allows for a faithful restoration of the object's structured motion.\lyj{Comments: Merge into one Table.}}
  \label{tab:motion_geo_tracking} 
  \vspace{-0.25cm}
  \resizebox{\textwidth}{!}{%
  \begin{tabular}{
    cccccccccc
    }
    \toprule
    \multirow{2}{*}[-0.7ex]{\textbf{Method}} & \multicolumn{3}{c}{\textbf{Rubik's Cube}} & \multicolumn{3}{c}{\textbf{Treasure Box}} & \multicolumn{3}{c}{\textbf{Door}} \\
    \cmidrule(lr){2-4} \cmidrule(lr){5-7} \cmidrule(lr){8-10}
    & EPE \down & $\delta_{3D}^{.10}$ \up & $\delta_{3D}^{.05}$ \up & EPE \down & $\delta_{3D}^{.10}$ \up & $\delta_{3D}^{.05}$ \up & EPE \down & $\delta_{3D}^{.10}$ \up & $\delta_{3D}^{.05}$ \up \\
    \midrule
    Shape of Motion & 0.586 & 0.042 & 0.025 & 0.553 & 0.168 & 0.073 & 1.812 & 0.020 & 0.004 \\
    Ub4D        & 0.303 & 0.110 & 0.030 & 0.132 & 0.540 & 0.280 & 0.127 & 0.590 & 0.410 \\
    MovingParts & \cellcolor{second}0.174 & \cellcolor{third}0.682 & \cellcolor{second}0.637 & 0.168 & 0.542 & 0.243 & \cellcolor{second}0.059 & \cellcolor{second}0.833 & \cellcolor{third}0.605 \\
    SP-GS & \cellcolor{third}0.179 & \cellcolor{second}0.691 & \cellcolor{third}0.623 & \cellcolor{third}0.078 & \cellcolor{second}0.789 & \cellcolor{third}0.622 & \cellcolor{third}0.066 & 0.737 & 0.558 \\ 
    DG-Mesh & 0.181 & 0.671 & 0.616 & \cellcolor{second}0.069 & \cellcolor{third}0.761 & \cellcolor{second}0.672 & 0.067 & \cellcolor{third}0.774 & \cellcolor{second}0.629 \\
    \midrule
    \textbf{Ours} & \cellcolor{best}0.063 & \cellcolor{best}0.877 & \cellcolor{best}0.869 & \cellcolor{best}0.006 & \cellcolor{best}1.000 & \cellcolor{best}0.998 & \cellcolor{best}0.011 & \cellcolor{best}0.990 & \cellcolor{best}0.957 \\
    \bottomrule
  \end{tabular}%
  } 
  \vspace{4mm} 
  \resizebox{\textwidth}{!}{%
  \begin{tabular}{
    cccccccccc
    }
    \toprule
    \multirow{2}{*}[-0.7ex]{\textbf{Method}} & \multicolumn{3}{c}{\textbf{Pliers}} & \multicolumn{3}{c}{\textbf{Folding Chair}} & \multicolumn{3}{c}{\textbf{Sunglasses}} \\
    \cmidrule(lr){2-4} \cmidrule(lr){5-7} \cmidrule(lr){8-10}
    & EPE \down & $\delta_{3D}^{.10}$ \up & $\delta_{3D}^{.05}$ \up & EPE \down & $\delta_{3D}^{.10}$ \up & $\delta_{3D}^{.05}$ \up & EPE \down & $\delta_{3D}^{.10}$ \up & $\delta_{3D}^{.05}$ \up \\
    \midrule
    Shape of Motion & 1.305 & 0.215 & 0.139 & 0.656 & 0.357 & 0.185 & 1.383 & 0.258 & 0.100 \\
    Ub4D        & 0.177 & 0.180 & 0.070 & 0.160 & 0.230 & 0.070 & 0.173 & 0.260 & 0.070 \\
    MovingParts & 0.118 & 0.416 & 0.245 & 0.083 & 0.747 & 0.560 & \cellcolor{third}0.060 & \cellcolor{third}0.858 & \cellcolor{third}0.654 \\
    SP-GS & \cellcolor{third}0.050 & \cellcolor{third}0.870 & \cellcolor{third}0.678 & \cellcolor{third}0.071 & \cellcolor{third}0.752 & \cellcolor{third}0.593 & 0.085 & 0.748 & 0.362 \\
    DG-Mesh & \cellcolor{best}0.010 & \cellcolor{best}0.999 & \cellcolor{best}0.975 & \cellcolor{second}0.037 & \cellcolor{second}0.898 & \cellcolor{second}0.717 & \cellcolor{second}0.015 & \cellcolor{second}0.969 & \cellcolor{second}0.941 \\
    \midrule
    \textbf{Ours} & \cellcolor{second}0.014 & \cellcolor{second}0.993 & \cellcolor{second}0.950 & \cellcolor{best}0.017 & \cellcolor{best}0.953 & \cellcolor{best}0.936 & \cellcolor{best}0.007 & \cellcolor{best}1.000 & \cellcolor{best}0.991 \\
    \bottomrule
  \end{tabular}%
  } 
  \vspace{-0.5cm}
\end{table*}

\subsection{Dynamic Primitive Optimization}
\label{sec:optim}
\subsubsection{Initialization}
To ensure proper gradient propagation to the primitive parameters, we must define a bounded motion range at initialization. This range is provided by the dataset. We then randomly sample $K$ points within this given range as the initial canonical $T$ values for the $K$ primitives. $T,R,\epsilon,s$ are all initialized randomly.

\subsubsection{Loss Functions}
Our loss function includes the rendering loss $\mathcal{L}_{\text{gs}}$ from 3DGS and a mask loss $\mathcal{L}_{\text{mask}}$. The mask loss is computed between the ground truth mask and the mask obtained from the primitive mesh rasterization. The final loss function is defined as:
\begin{equation}
\begin{aligned}
    \mathcal{L} ={} & \mathcal{L}_{\text{gs}} + \mathcal{L}_{\text{mask}} + \lambda_{\text{over}}\mathcal{L}_{\text{over}} \\
                 & + \lambda_{\text{parsi}}\mathcal{L}_{\text{parsi}} + \lambda_{\text{vol}}\mathcal{L}_{\text{vol}} + \mathcal{L}_{\text{deform}}.
\end{aligned}
\end{equation}
We retain two regularization terms from previous work \cite{monnier2023differentiable}: $\mathcal{L}_{\text{over}}$ to prevent excessive primitive overlap, and $\mathcal{L}_{\text{parsi}}$ to reduce the opacity of redundant primitives, facilitating their removal. $\mathcal{L}_{\text{vol}}$ is a penalty term for primitives with excessively small volumes, computed as:
\begin{equation}
\label{equ:volume}
\begin{aligned}
V &= 4 s_1 s_2 s_3 \epsilon_1 \cdot \left( \frac{\Gamma\left(1 + \frac{\epsilon_2}{2}\right)^2}{\Gamma(1 + \epsilon_2)} \right) \cdot \mathrm{B}\left(\frac{\epsilon_1}{2}, \epsilon_1 + 1\right), \\
\mathcal{L}_{\text{vol}} &= \frac{1}{V},
\end{aligned}
\end{equation}
where $\Gamma$ is the Gamma function and $\mathrm{B}$ is the Beta function. This term encourages primitives to increase in volume early in optimization, thereby reducing redundant primitives. $\mathcal{L}_{\text{deform}}$ consists of several regularization terms related to motion:
\begin{equation}
    \mathcal{L}_{\text{deform}} = \lambda_{\text{smooth}}\mathcal{L}_{\text{smooth}}+\lambda_{\text{trans}}\mathcal{L}_{\text{trans}}+\lambda_{\text{back}}\mathcal{L}_{\text{back}}.
\end{equation}
The purpose of $\mathcal{L}_{\text{smooth}}$ is to stabilize the deformation process and prevent abrupt deformation changes:
\begin{equation}
\begin{aligned}
    a_t &= \Delta T_t - \Delta T_{t-1},\\
    \delta_t &= \text{rot}(\Delta R_{t}) \cdot \text{rot}(\Delta R_{t-1})^{-1}, \\
    \mathcal{L}_{\text{smooth}} &= (a_{t+1} - a_t) + \text{log}_{SO3}[\delta_{t+1} \cdot \delta_t^{-1}].
\end{aligned}
\end{equation}
$\mathcal{L}_{\text{trans}} = \frac{1}{3}\sum |\Delta T|$ penalizes excessive primitive motion early in training, preventing them from moving out of bounds.
Both $\mathcal{L}_{\text{smooth}}$ and $\mathcal{L}_{\text{trans}}$ aim to improve training stability.
$\mathcal{L}_{\text{back}} = \frac{1}{3}\sum |\Delta T_{\text{forw}}-\Delta T_{\text{backw}}|$ supervises the inverse deformation network, ensuring that positional changes made to primitives in the current space can be mapped back to the canonical space.

\subsubsection{Adaptive Control for Dynamic Primitives}
The simplistic primitive count control of previous methods \cite{monnier2023differentiable,gao2025self} causes training in dynamic scenes to be highly initialization-dependent and unstable. Therefore, it is crucial to adaptively adjust the primitive count and distribution after initialization. Inspired by the adaptive control in Gaussian \cite{kerbl20233d} methods, we propose a tailored strategy for dynamic primitives.\chong{remove the word "similar", this makes your contribution seems less novel...} At regular iteration intervals, we apply the following three operations to the primitive set.

\textbf{Clone operation.} 
We monitor the gradients of the 3D Gaussians on each primitive. We track the proportion of these Gaussians whose gradient exceeds a threshold $\tau_g = 4e^{-5}$. If this proportion exceeds a threshold $\tau_p = 0.15$, the primitive is marked for cloning. This operation allows the primitive to better occupy the object's region. If the primitive's scale is excessively large, we first reduce its scale and then clone it. Otherwise, we clone an identical primitive in the same position.

\textbf{Merge operation.}
Our principle is to achieve high representational completeness with the fewest primitives possible. Therefore, we apply a merge operation to primitives that have a high degree of overlap. We first compute the mutual overlap ratio between primitives, which is calculated by averaging the overlap across all timestamps.
We then group primitives by constructing a graph. An edge is added from a primitive only to its highest-overlapping neighbor, provided the overlap exceeds a threshold $\tau_o$, detailed in Sec. \ref{sec:ablation}. 
Primitives lacking significant overlap remain isolated, with final groups defined by the graph's connected components.
We then perform the merge operation based on these groups. Groups containing only one primitive remain unchanged. For groups with multiple primitives, we first apply a pruning step where primitives are removed if their volume (using Equ. \ref{equ:volume}) is less than one-third of the group's maximum volume, or if their overlap with another group member exceeds $80\%$. The remaining primitives are then merged into a new primitive. Its $T$ value is the volume-weighted average of the remaining primitives' $T$ values, while all other attributes are adopted from the largest primitive in the group. Finally, we use the inverse deformation network to map the new primitive back to the canonical space.

\textbf{Prune operation.}
We remove primitives whose opacity is below a threshold $\tau_{\alpha}=0.3$.\chong{=?, or are you using a different threshold for each object?} We also prune primitives based on volume. After sorting all volumes, if a jump greater than $10\times$ is found between adjacent values, all primitives smaller than that jump's lower bound are removed.

\subsection{Post-Training Refinement}
\label{sec:refine}
After the main training, we employ a refinement stage to further optimize primitive motion and appearance. In this stage, we fine-tune the model by lowering all learning rates. We freeze the canonical $T$ and $R$ parameters for all primitives, while unfreezing all parameters of the bound 3D Gaussians. Specifically, their barycentric coordinates are now optimized, though the centers remain constrained to the primitive surfaces. The number of primitives is also fixed. This refinement process yields more accurate motion reconstruction and higher-quality appearance and geometry.
\section{Experiments}
\subsection{Datasets}
To validate our method's capability for dynamic modeling of structured objects, we have rendered a synthetic dataset named Dynamic Primitive Dataset. It contains 6 structured objects based on assets from PartNet-Mobility Dataset \cite{Xiang_2020_SAPIEN}, including an openable Treasure Box, a swinging Door, moving Pliers, a movable Folding Chair, foldable Sunglasses, and a rotating Rubik’s Cube which we custom-build using SAPIEN \cite{Xiang_2020_SAPIEN}. 
Each sequence is monocular, with a camera trajectory mimicking a practical real-world capture setup on the upper hemisphere, and provides ground truth dynamic meshes for each timestamp.\chong{is camera motion continuous? otherwise it might be not that monocular, or in some sense, too far from a real capturing setup} 
We also test our method on selected sequences from D-NeRF Dataset \cite{pumarola2021d} to evaluate its effectiveness in general cases.

\subsection{Implementation Details}
Our implementation uses 3DGS rendering pipeline \cite{kerbl20233d} and Nvdiffrast \cite{Laine2020diffrast} for mask rasterization. We use Adam \cite{adam2014method} optimizer with a learning rate of 0.001 for primitive parameters, which is scaled by 0.01 in the refinement stage. Both main training and refinement run for 60k iterations on one NVIDIA 3090 GPU, with primitive adaptive control applied every 2k iterations. Due to compositional differences, initial primitive counts vary by object type, as detailed in supplementary. The loss weights are $\lambda_{\text{over}}=1$, $\lambda_{\text{parsi}}=0.1$, $\lambda_{\text{vol}}=0.05$, $\lambda_{\text{smooth}}=1$, $\lambda_{\text{trans}}=0.5$, and $\lambda_{\text{back}}=0.01$.

\subsection{Evaluation of Structured Motion Accuracy}

\begin{table*}[!t] 
  \centering
  \scriptsize
  \caption{\textbf{Structured Dynamic Reconstruction Results on Dynamic Primitive Dataset.} The unit for CD$_d$ is ${10}^{-3}$. 
  For a fair comparison, we select other dynamic geometry-based methods. Despite our structured primitive representation loses some geometric detail, it still achieves strong quantitative results. Our approach shows excellent results especially in cases with large motion, such as Rubik's Cube.}
  \label{tab:motion_geo_mesh}
  \resizebox{\textwidth}{!}{
  \begin{tabular}{ccccccccccccc}
    \toprule
    \multirow{2}{*}[-0.7ex]{\textbf{Method}} & \multicolumn{2}{c}{\textbf{Rubik's Cube}} & \multicolumn{2}{c}{\textbf{Treasure Box}} & \multicolumn{2}{c}{\textbf{Door}} & \multicolumn{2}{c}{\textbf{Pliers}} & \multicolumn{2}{c}{\textbf{Folding Chair}} & \multicolumn{2}{c}{\textbf{Sunglasses}} \\
    \cmidrule(lr){2-3} \cmidrule(lr){4-5} \cmidrule(lr){6-7} \cmidrule(lr){8-9} \cmidrule(lr){10-11} \cmidrule(lr){12-13}
    & CD$_d$ \down & EMD$_d$ \down & CD$_d$ \down & EMD$_d$ \down & CD$_d$ \down & EMD$_d$ \down & CD$_d$ \down & EMD$_d$ \down & CD$_d$ \down & EMD$_d$ \down & CD$_d$ \down & EMD$_d$ \down \\
    \midrule
    Ub4D        & \cellcolor{second}6.505 & 0.303 & 9.732 & \cellcolor{second}0.178 & 81.168 & 0.250 & 9.655 & 0.154 & 4.649 & 0.251 & 19.052 & 0.209 \\
    DG-Mesh     & 9.873 & \cellcolor{second}0.206 & \cellcolor{second}4.737 & 0.182 & \cellcolor{second}4.940 & \cellcolor{second}0.197 & \cellcolor{second}0.544 & \cellcolor{second}0.075 & \cellcolor{second}2.606 & \cellcolor{best}0.125 & \cellcolor{second}4.485 & \cellcolor{second}0.103 \\
    \midrule
    \textbf{Ours} & \cellcolor{best}3.237 & \cellcolor{best}0.114 & \cellcolor{best}1.848 & \cellcolor{best}0.118 & \cellcolor{best}2.777 & \cellcolor{best}0.194 & \cellcolor{best}0.541 & \cellcolor{best}0.058 & \cellcolor{best}2.483 & \cellcolor{second}0.146 & \cellcolor{best}3.773 & \cellcolor{best}0.096 \\
    \bottomrule
  \end{tabular}
  }
\end{table*}

\begin{table*}[!t] 
  \centering
  \caption{\textbf{General Dynamic Rendering Results.} We choose humanoid-like cases from D-NeRF Dataset for testing our method on more general scenes. 
  Following DG-Mesh, we benchmark our method against the geometry-based results of other approaches.
  Despite some loss in geometric detail, our method still achieves competitive results. Moreover, our approach provides a structured geometric representation, which other methods cannot offer.}
  \label{tab:d_nerf_rendering}
  \resizebox{\textwidth}{!}{%
  \begin{tabular}{cccccccccccccccc}
    \toprule
    \multirow{2}{*}[-0.7ex]{\textbf{Method}} & \multicolumn{3}{c}{\textbf{Jumpingjacks}} & \multicolumn{3}{c}{\textbf{Hook}} & \multicolumn{3}{c}{\textbf{Hellwarrior}} & \multicolumn{3}{c}{\textbf{Mutant}} & \multicolumn{3}{c}{\textbf{Standup}} \\
    \cmidrule(lr){2-4} \cmidrule(lr){5-7} \cmidrule(lr){8-10} \cmidrule(lr){11-13} \cmidrule(lr){14-16}
    & PSNR \up & SSIM \up & LPIPS \down & PSNR \up & SSIM \up & LPIPS \down & PSNR \up & SSIM \up & LPIPS \down & PSNR \up & SSIM \up & LPIPS \down & PSNR \up & SSIM \up & LPIPS \down \\
    \midrule
    D-NeRF      & 22.255 & 0.914 & 0.103 & 20.300 & 0.889 & 0.108 & 18.907 & 0.877 & 0.129 & 21.070 & 0.906 & 0.077 & 23.380 & 0.925 & 0.069 \\
    K-Plane     & 25.240 & 0.937 & 0.068 & 22.503 & 0.900 & 0.094 & 18.073 & 0.881 & 0.123 & 23.226 & 0.923 & 0.064 & 25.778 & 0.946 & \cellcolor{third}0.048 \\
    HexPlane    & \cellcolor{third}27.078 & \cellcolor{third}0.954 & \cellcolor{third}0.052 & \cellcolor{third}24.513 & \cellcolor{third}0.929 & \cellcolor{second}0.070 & \cellcolor{third}21.250 & \cellcolor{third}0.917 & \cellcolor{second}0.094 & \cellcolor{second}26.811 & \cellcolor{second}0.953 & \cellcolor{best}0.045 & \cellcolor{second}27.931 & \cellcolor{second}0.965 & \cellcolor{second}0.035 \\
    TiNeuVox-B  & 23.621 & 0.932 & 0.075 & 21.429 & 0.908 & 0.085 & 18.657 & 0.883 & 0.118 & 22.967 & 0.925 & 0.064 & 24.263 & 0.941 & 0.051 \\
    DG-Mesh     & \cellcolor{best}31.769 & \cellcolor{best}0.977 & \cellcolor{second}0.045 & \cellcolor{best}27.884 & \cellcolor{best}0.954 & \cellcolor{third}0.074 & \cellcolor{best}25.460 & \cellcolor{best}0.959 & \cellcolor{best}0.084 & \cellcolor{best}30.400 & \cellcolor{best}0.968 & \cellcolor{second}0.055 & \cellcolor{best}30.208 & \cellcolor{best}0.974 & 0.051 \\
    \midrule
    \textbf{Ours} & \cellcolor{second}29.069 & \cellcolor{second}0.965 & \cellcolor{best}0.034 & \cellcolor{second}25.656 & \cellcolor{second}0.935 & \cellcolor{best}0.053 & \cellcolor{second}22.336 & \cellcolor{second}0.936 & \cellcolor{best}0.084 & \cellcolor{third}26.518 & \cellcolor{third}0.945 & \cellcolor{best}0.045 & \cellcolor{third}27.651 & \cellcolor{third}0.963 & \cellcolor{best}0.032 \\
    \bottomrule
  \end{tabular}%
  }
\end{table*}

Previous methods for dynamic scene reconstruction simply extended traditional metrics like Chamfer Distance (CD) and Earth Mover's Distance (EMD) over time \cite{liu2024dynamic,johnson2022ub4d}. However, this approach merely compares per-frame meshes and thus cannot accurately reflect motion modeling quality, as it fails to evaluate if predicted geometry parts move similarly to their corresponding ground truth parts. A metric is needed to quantify this similarity.
Therefore, we propose a new metric: \textbf{structured motion tracking accuracy}.\chong{maybe highlight with bold text} We first sample $N$ points on a canonical ground truth mesh and compute their corresponding positions across all timestamps,
which serve as ground truth tracking values. Next, we use the method under evaluation to predict the motion of these $N$ points. Then, we compute the difference using 3D End-Point Error (EPE) and the percentage of points within 0.05 ($\delta_{3D}^{.05}$) and 0.10 ($\delta_{3D}^{.10}$) metric scale thresholds, similar to 3D tracking tasks \cite{wang2025shape}. This metric directly reflects how well a method models dynamic geometry motion.

Tab. \ref{tab:motion_geo_tracking} shows our motion tracking results on Dynamic Primitive Dataset. For a comprehensive comparison, we select monocular methods that produce geometry (DG-Mesh \cite{liu2024dynamic}, Ub4D \cite{johnson2022ub4d}) and monocular NVS methods that only provide tracking (MovingParts \cite{yang2023movingparts}, Shape of Motion \cite{wang2025shape}, SP-GS \cite{icml2024-sp-gs}). The results show our method significantly improves motion modeling accuracy, especially on large-scale, long-duration motions where other methods degrade. We also evaluate the predicted dynamic consistent meshes using traditional metrics CD$_d$ and EMD$_d$ in dynamic settings. 
As shown in Tab. \ref{tab:motion_geo_mesh}, our method achieves strong geometric accuracy despite prioritizing structured representation over fine-grained detail.
Fig. \ref{fig:mesh_vis} provides a visual comparison, highlighting our method's stable and correct results in large-motion cases such as the 360-degree rotation of a Rubik's Cube part and the 180-degree rotation of a Treasure Box lid, where other methods fail in both motion and geometry reconstruction.

\subsection{Evaluation on General Dynamic Cases}
\begin{figure}[!t]
  \centering
   \includegraphics[width=1\linewidth]{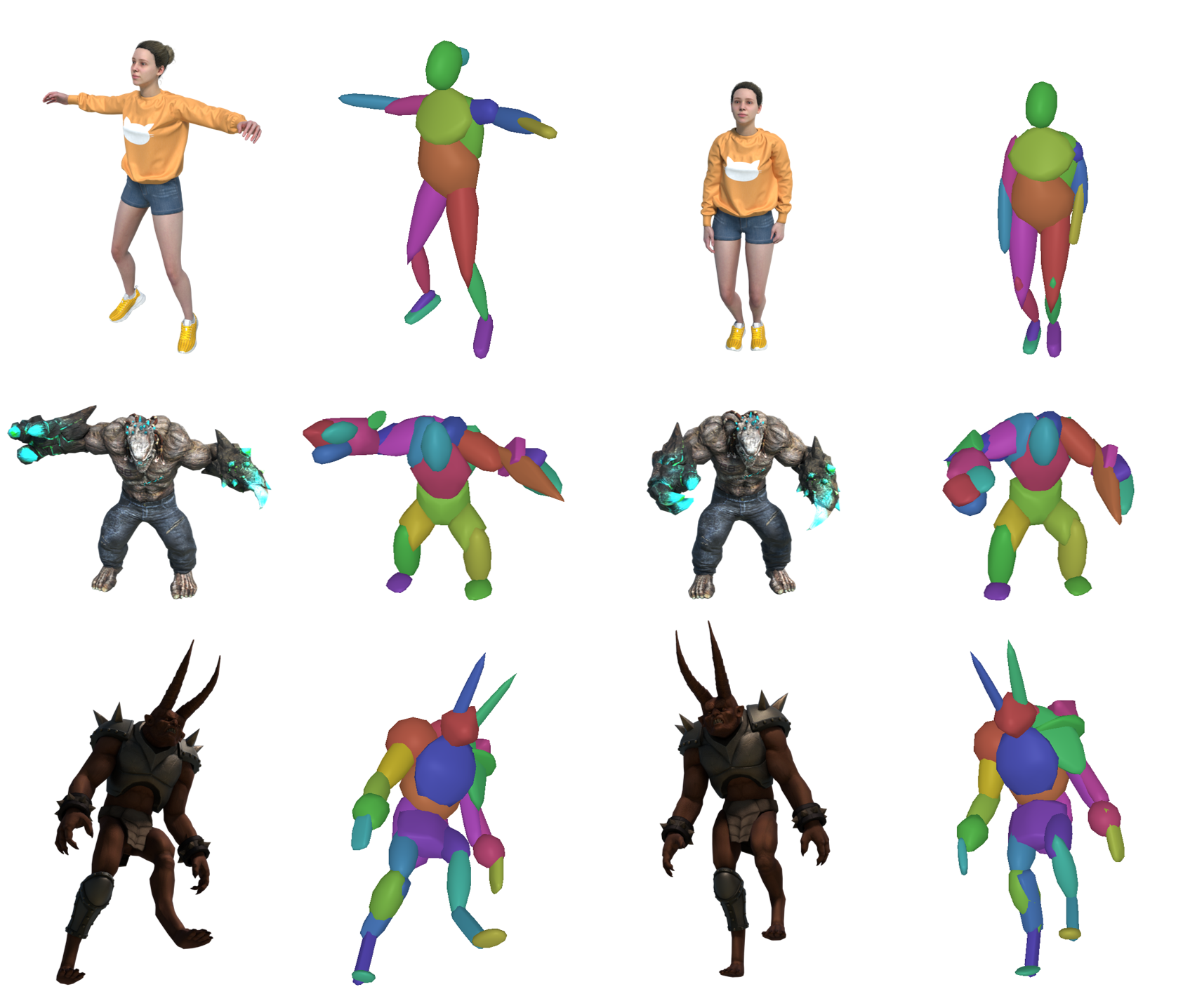}
   \caption{\textbf{Visualization of Structured Dynamic Modeling in D-NeRF Dataset.} 
   Our method also provides good structured geometry reconstruction results in general cases.
    }
   \label{fig:dnerf_vis}
   \vspace{-0.25cm}
\end{figure}

To validate our method's performance in more general scenes, we also test it on D-NeRF Dataset \cite{pumarola2021d}. We select the humanoid-like cases for testing, evaluating the rendered images using PSNR, SSIM, and LPIPS. 
Following \cite{liu2024dynamic}, we maintain the same settings and compare the geometry-based rendering quality against methods \cite{pumarola2021d,fridovich2023k,cao2023hexplane,fang2022fast}, as detailed in Tab. \ref{tab:d_nerf_rendering}.
Our method achieves competitive rendering quality while also providing a structured geometric representation. Fig. \ref{fig:dnerf_vis} visualizes the structured geometric results for some cases. We also test our method on real-world cases, as detailed in supplementary.

\subsection{Ablation Studies}
\label{sec:ablation}
In this section, we analyze our framework's core designs. As shown in Tab. \ref{tab:ablation_deform}, we first validate the necessity of using deformation networks to model motion. The motion tracking results show that the deformation network vastly outperforms directly optimizing primitive poses, showing it better models the object's true motion. Accurate motion modeling in turn improves geometry quality, as reflected in the dynamic reconstruction metrics.

\begin{table}[!t] 
  \centering 
  \caption{\textbf{Ablation study for Motion Modeling.} 
  We use Rubik’s Cube from Dynamic Primitive Dataset to demonstrate that a deformation network significantly improves motion modeling accuracy, which is superior to creating a per-frame learnable pose transformation for each primitive.
  }
  \label{tab:ablation_deform}
  \resizebox{\columnwidth}{!}{%
  \begin{tabular}{cccccc}
    \toprule
    \textbf{Method} & EPE \down & $\delta_{3D}^{.05}$ \up & $\delta_{3D}^{.10}$ \up & CD$_d$ \down & EMD$_d$ \down \\
    \midrule
    w/o. deform        & 0.177 & 0.692 & 0.632 & 3.642 & 0.168 \\
    w. deform & \textbf{0.063} & \textbf{0.877} & \textbf{0.869} & \textbf{3.237} & \textbf{0.114} \\
    \bottomrule
  \end{tabular}
  }
\end{table}

\begin{figure}[!t]
  \centering
   \includegraphics[width=1\linewidth]{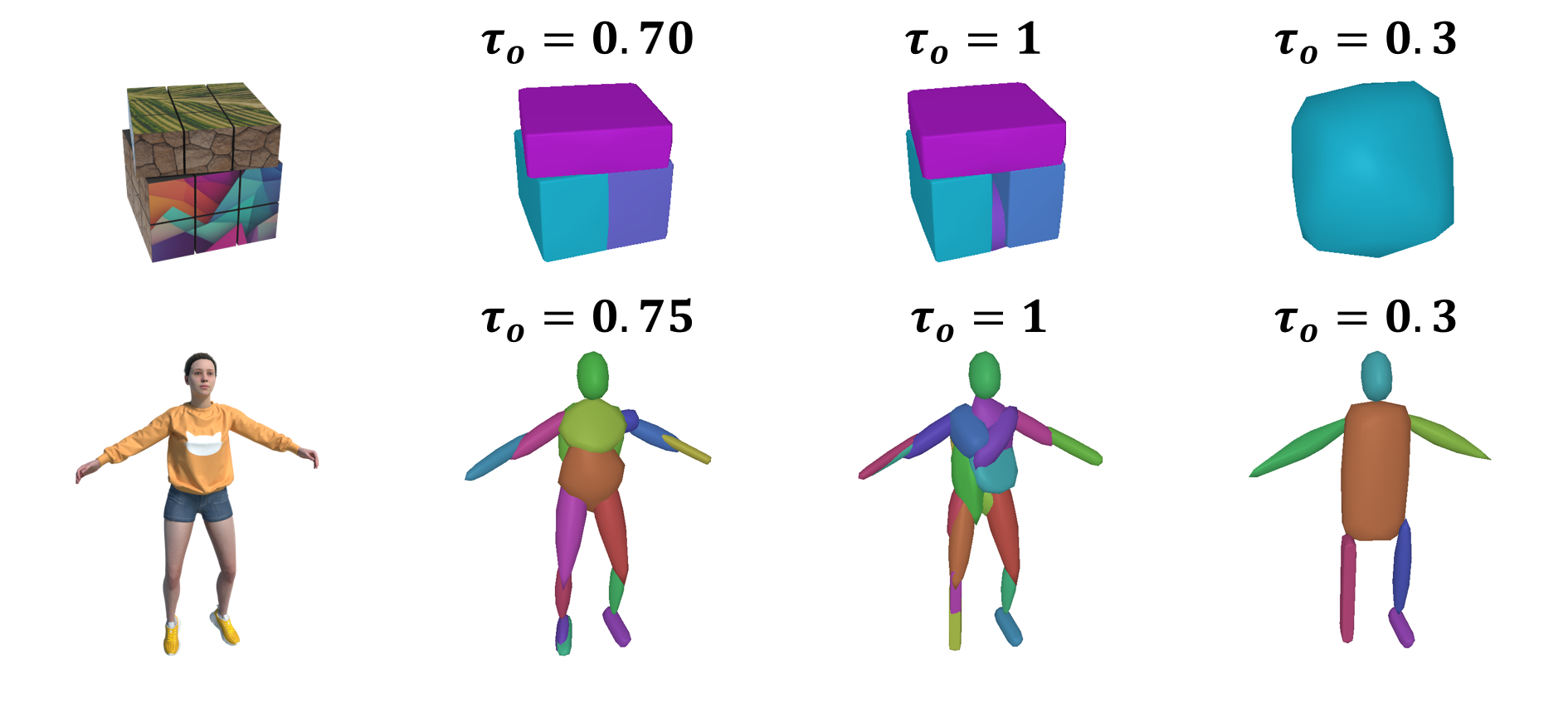}
   \caption{\textbf{Ablation Study for Primitive Adaptive Control's Merge Operation.} 
   The leftmost image is GT. The first column shows the default result. A high threshold $\tau_{o}$ causes redundancy, such as extra primitives in the cube's center or excessive primitives on the torso. A low $\tau_{o}$ yields poor representation, degenerating the cube into one primitive and over-simplifying the torso.
    }
   \label{fig:ablation_comb}
   \vspace{-0.25cm}
\end{figure}

Subsequently, we validate the effectiveness of our dynamic primitive adaptive control strategy. Besides a basic prune operation, our strategy also includes two more effective operations of merging and cloning. The merge operation aims to reduce primitive redundancy, which is difficult to demonstrate quantitatively. Therefore, we use the visualizations in Fig. \ref{fig:ablation_comb} to intuitively show the effects of the merge operation and its related threshold $\tau_o$. We conduct experiments on both structured and humanoid objects. Adjusting $\tau_o$ controls the merge intensity. A low intensity ($\tau_o=1$) merges only fully overlapped primitives, while a high intensity ($\tau_o=0.3$) merges primitives at $30\%$ overlap. Our preset suitable range is $\tau_o=0.7 \pm 0.05$. Fig. \ref{fig:ablation_comb} clearly shows that low merge intensity leads to significant redundancy, with unnecessary primitives in many regions. Conversely, excessive merge intensity has a more severe impact, directly reducing the framework's expressiveness and causing a loss of necessary object structures. The merge operation with appropriate intensity effectively enhances the compactness and interpretability of the primitive set while preserving modeling accuracy.

The clone operation adds new primitives to enhance the set's potential expressiveness when the current representation is insufficient. Its effect is less obvious on structured objects but is highly effective in general scenes, as the initial primitive set there often fails to capture all object details. Therefore, we select sequences from the D-NeRF dataset to demonstrate the necessity of cloning.
As shown in Tab. \ref{tab:ablation_clone}, in these cases involving detailed and complex shapes, cloning effectively improves primitive expressiveness, which is directly reflected in the rendering results. Without cloning, the object lacks suitable primitives in many regions, leading to poorer representation and lower rendering quality. The clone operation enables primitives to achieve better performance in more diverse scenes.

\begin{table}[!t] 
  \centering 
  \caption{
  \textbf{Ablation Study for Primitive Adaptive Control's Clone Operation.} 
  In cases with relatively complex shape details, cloning improves representing ability by adaptively adding new primitives where details are lacking. Results on D-NeRF Dataset confirm that it improves our dynamic modeling capability.
  }
  \label{tab:ablation_clone}
  \resizebox{\columnwidth}{!}{%
  \begin{tabular}{ccccccc}
    \toprule
    \multirow{2}{*}[-0.7ex]{\textbf{Method}} & \multicolumn{3}{c}{\textbf{Jumpingjacks}} & \multicolumn{3}{c}{\textbf{Mutant}} \\
    \cmidrule(lr){2-4} \cmidrule(lr){5-7}
    & PSNR \up & SSIM \up & LPIPS \down & PSNR \up & SSIM \up & LPIPS \down \\
    \midrule
    w/o. clone        & 25.873 & 0.956 & 0.058 & 24.667 & 0.934 & 0.062 \\
    w. clone & \textbf{29.069} & \textbf{0.965} & \textbf{0.034} & \textbf{26.518} & \textbf{0.945} & \textbf{0.045} \\
    \bottomrule
  \end{tabular}
  }
\end{table}

\subsection{Applications}
Our structured representation enables several downstream applications. After obtaining the structured representation from our method, part-level editing can be easily performed, such as adding, deleting, or scaling specific components. Furthermore, articulation of the primitives can be easily performed in software such as Blender to assign novel motions to the object. We provide specific demonstrations of these applications in our supplementary video.
\section{Conclusions}
In this work, we introduce D-Prism, a novel framework that uses structured primitives for dynamic geometry reconstruction, providing part-based representation and superior motion modeling for various types of objects. To validate our approach, we present the Dynamic Primitive Dataset and a new structured motion tracking metric. Experiments confirm our method's ability to accurately reconstruct both object motion and structured geometry.
Despite these results, our method exhibits some sensitivity to primitive initialization and struggles with complex topologies like a torus or dense and slender structures. Future work will focus on expanding primitive expressiveness and learning hierarchical, skeleton-like relationships between primitives through training. We also plan to assess our method's potential for evaluating the structural consistency of generated videos, which could benefit embodied AI, world modeling and multimodal learning.

\section*{Acknowledgments}
This work was partially supported by NSF of China (No. 62425209). We thank Shang Liu and Yichen Shen for their assistance with illustration design. We also thank Jingjing Wang, Yuke Zhu, and Zhengdong Hong for their help with real-world scene testing.

{
    \small
    \bibliographystyle{ieeenat_fullname}
    \bibliography{main}
}

\clearpage
\setcounter{page}{1}

\maketitlesupplementary

\appendix
\section{More Details for Implementation}
\subsection{Primitive Initialization}
We provide detailed initialization parameters for all cases across different datasets to demonstrate our framework's reproducibility. These details are included here due to space constraints in the main text.
Here we list the initial primitive count $K$, the number of 3DGS allocated per primitive $N_{\text{gs}}$, and the scene scaling value $s_{\text{scene}}$ for different scenes. This $s_{\text{scene}}$ is multiplied by the primitive's scale ($s_1,s_2,s_3$) to ensure the primitive has a moderate initial size, helping it appear within the camera's view during the early stages of training, the same as \cite{monnier2023differentiable_supp}.
\begin{table}[H] 
  \centering 
  \caption{\textbf{Initialization Settings for Training.}}
  \label{tab:init_val}
  \resizebox{\columnwidth}{!}{%
  \begin{tabular}{ccccccc}
    \toprule
    \multirow{2}{*}[-0.7ex]{\textbf{Property}} & \multicolumn{6}{c}{\textbf{Dynamic Primitive Dataset}} \\
    \cmidrule(lr){2-7}
    & \textbf{Rubik’s Cube} & \textbf{Treasure Box} & \textbf{Door} & \textbf{Pliers} & \textbf{Folding Chair} & \textbf{Sunglasses} \\
    \midrule
    $K$                & 10 & 10 & 20 & 50 & 50 & 25 \\
    $N_{\text{gs}}$    & 50k & 50k & 10k & 8k & 10k & 10k \\
    $s_{\text{scene}}$ & 0.4 & 0.4 & 0.2 & 0.1 & 0.1 & 0.25 \\
    \bottomrule
  \end{tabular}%
  } 
  
  \vspace{1mm} 

  \resizebox{\columnwidth}{!}{%
  \scriptsize
  \begin{tabular}{cccccc}
    \toprule
    \multirow{2}{*}[-0.7ex]{\textbf{Property}} & \multicolumn{5}{c}{\textbf{D-NeRF Dataset}} \\
    \cmidrule(lr){2-6}
    & \textbf{Jumpingjacks} & \textbf{Hook} & \textbf{Hellwarrior} & \textbf{Mutant} & \textbf{Standup} \\
    \midrule
    $K$             & 50 & 50 & 50 & 50 & 50 \\
    $N_{\text{gs}}$ & 5k & 5k & 5k & 5k & 5k \\
    $s_{\text{scene}}$ & 0.25 & 0.25 & 0.25 & 0.25 & 0.25 \\
    \bottomrule
  \end{tabular}%
  } 
\end{table}

\subsection{Details for Deformation Networks}
We begin with a 5k iteration warm-up stage, during which the deformation network is frozen. This stage focuses on initially optimizing primitive shapes and 3D Gaussian color parameters. We then unfreeze the network for the main training.

We implement both the forward and backward deformation networks as Multi-Layer Perceptrons (MLPs). The architecture for both comprises $D=8$ layers, each with a hidden dimensionality of $W=256$. To help capture high-frequency variations, the network inputs, the primitive's motion translation $T$ and the temporal label $t$, are first mapped into a higher-dimensional Fourier space using a positional encoding function $\xi_k(p)$. We set the encoding parameter $k=10$ for $T$ and $k=6$ for $t$.

The network culminates in a single linear layer, without an activation function, which predicts the offsets for the primitive parameters: translation ($\Delta T$) and rotation ($\Delta R$). These predicted offsets allow for the dynamic adjustment of the primitives. At initialization, we set the final layer's weights for $\Delta T$ to zero to prevent large initial displacements from the deformation.

\subsection{Details for Computation Resource}
Our method's training time varies across scenes. On a single NVIDIA 3090 GPU, the main training process takes 2-3 hours, and the refinement stage takes 1-2 hours. The VRAM requirement of GPU also fluctuates, typically ranging from 8G to 16G, depending on scene complexity. For objects with more detail, our primitive adaptive control strategy generates more primitives, which increases the peak GPU memory usage during training.

\section{Visualization Results for Dynamic Primitive Dataset}
\begin{table*}[!t] 
  \centering
  \scriptsize
  \caption{\textbf{Structured Motion Rendering Results on Dynamic Primitive Dataset.} We select dedicated test images and compute the average results across all of them. 
  For a fair comparison, we select other geometry-based methods and using DG-Mesh*, which are its gaussian rendering results.
  Our structured representation simplifies the scene, which leads to some loss of detail. Despite this, our method maintains good rendering metrics. Our approach also shows excellent results in cases with large motion, such as Rubik's Cube.}
  \label{tab:motion_geo_rendering}
  \resizebox{\textwidth}{!}{
  \begin{tabular}{ccccccccccccc}
    \toprule
    \multirow{2}{*}[-0.7ex]{\textbf{Method}} & \multicolumn{2}{c}{\textbf{Rubik's Cube}} & \multicolumn{2}{c}{\textbf{Treasure Box}} & \multicolumn{2}{c}{\textbf{Door}} & \multicolumn{2}{c}{\textbf{Pliers}} & \multicolumn{2}{c}{\textbf{Folding Chair}} & \multicolumn{2}{c}{\textbf{Sunglasses}} \\
    \cmidrule(lr){2-3} \cmidrule(lr){4-5} \cmidrule(lr){6-7} \cmidrule(lr){8-9} \cmidrule(lr){10-11} \cmidrule(lr){12-13}
    & PSNR \up & SSIM \up & PSNR \up & SSIM \up & PSNR \up & SSIM \up & PSNR \up & SSIM \up & PSNR \up & SSIM \up & PSNR \up & SSIM \up \\
    \midrule
    Ub4D        & 17.199 & 0.848 & 20.858 & 0.919 & 21.877 & \cellcolor{second}0.964 & 23.520 & 0.982 & 17.707 & 0.946 & 17.907 & 0.983 \\
    DG-Mesh* & \cellcolor{second}22.540 & \cellcolor{second}0.913 & \cellcolor{second}22.110 & \cellcolor{second}0.935 & \cellcolor{best}26.821 & \cellcolor{best}0.973 & \cellcolor{second}31.436 & \cellcolor{best}0.989 & \cellcolor{second}27.200 & \cellcolor{best}0.980 & \cellcolor{best}33.063 & \cellcolor{best}0.989 \\
    \midrule
    \textbf{Ours} & \cellcolor{best}29.218 & \cellcolor{best}0.941 & \cellcolor{best}28.000 & \cellcolor{best}0.943 & \cellcolor{second}26.480 & \cellcolor{second}0.964 & \cellcolor{best}33.874 & \cellcolor{second}0.988 & \cellcolor{best}28.107 & \cellcolor{second}0.962 & \cellcolor{second}31.127 & 0.981 \\
    \bottomrule
  \end{tabular}
  }
\end{table*}

\begin{figure*}[!t]
  \centering
   \includegraphics[width=1\linewidth]{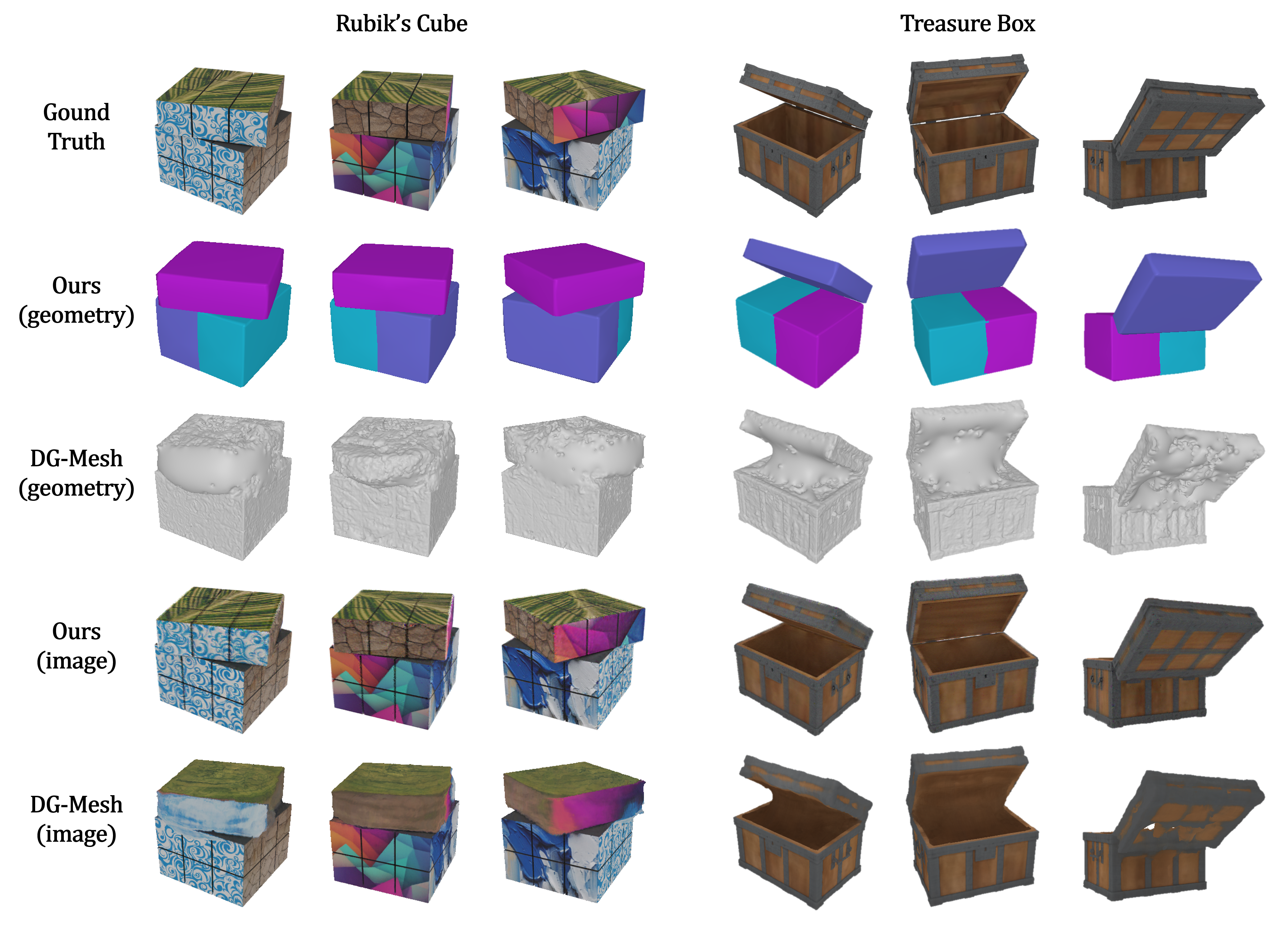}
   \caption{\textbf{Visualization Results for Dynamic Primitive Dataset.} We show both geometry results and rendering results.}
   \label{fig:pri_data_vis1}
\end{figure*}

\begin{figure*}[!t]
  \centering
   \includegraphics[width=1\linewidth]{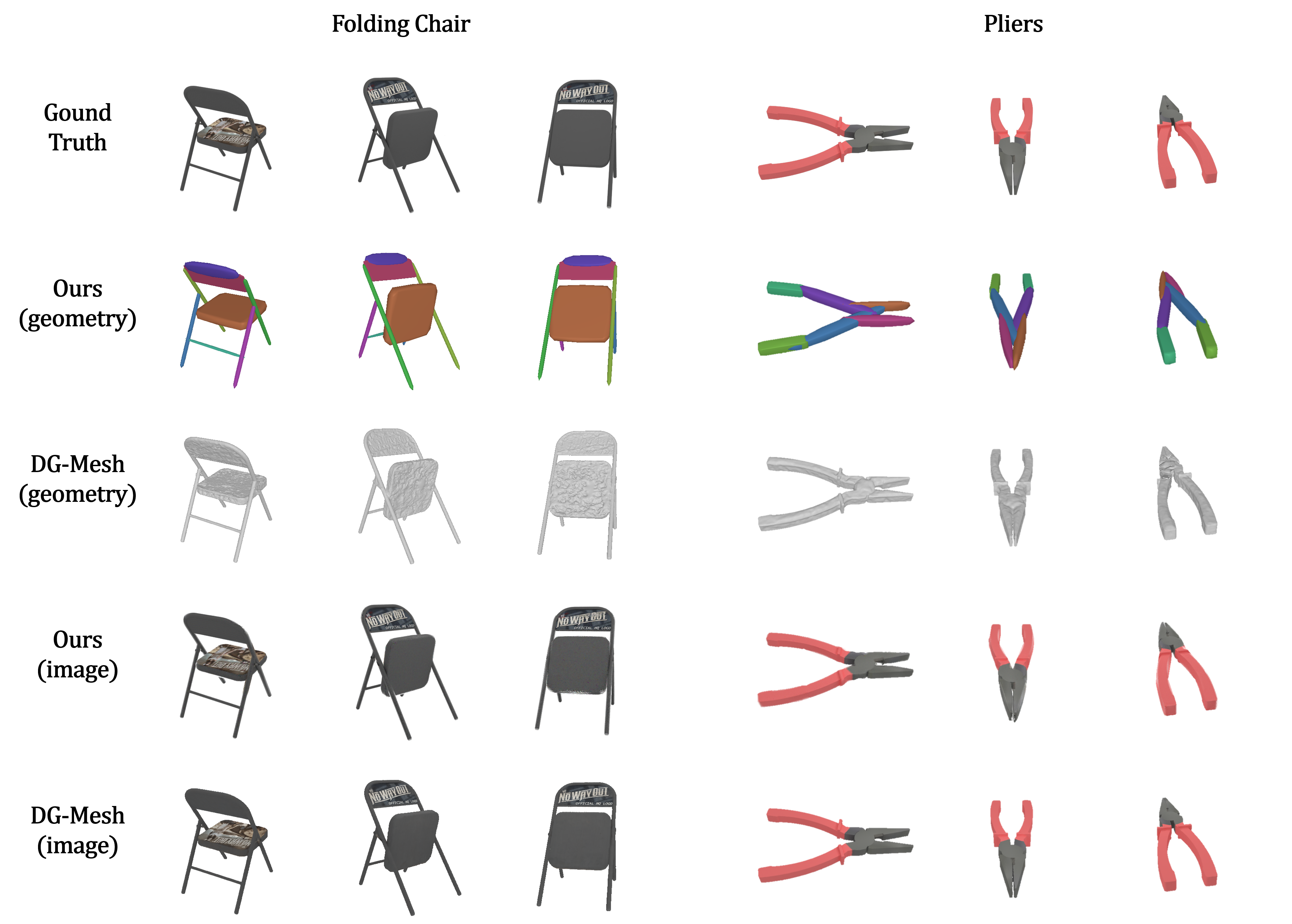}
   \caption{\textbf{Visualization Results for Dynamic Primitive Dataset.} We show both geometry results and rendering results.}
   \label{fig:pri_data_vis2}
\end{figure*}

\begin{figure*}[!t]
  \centering
   \includegraphics[width=1\linewidth]{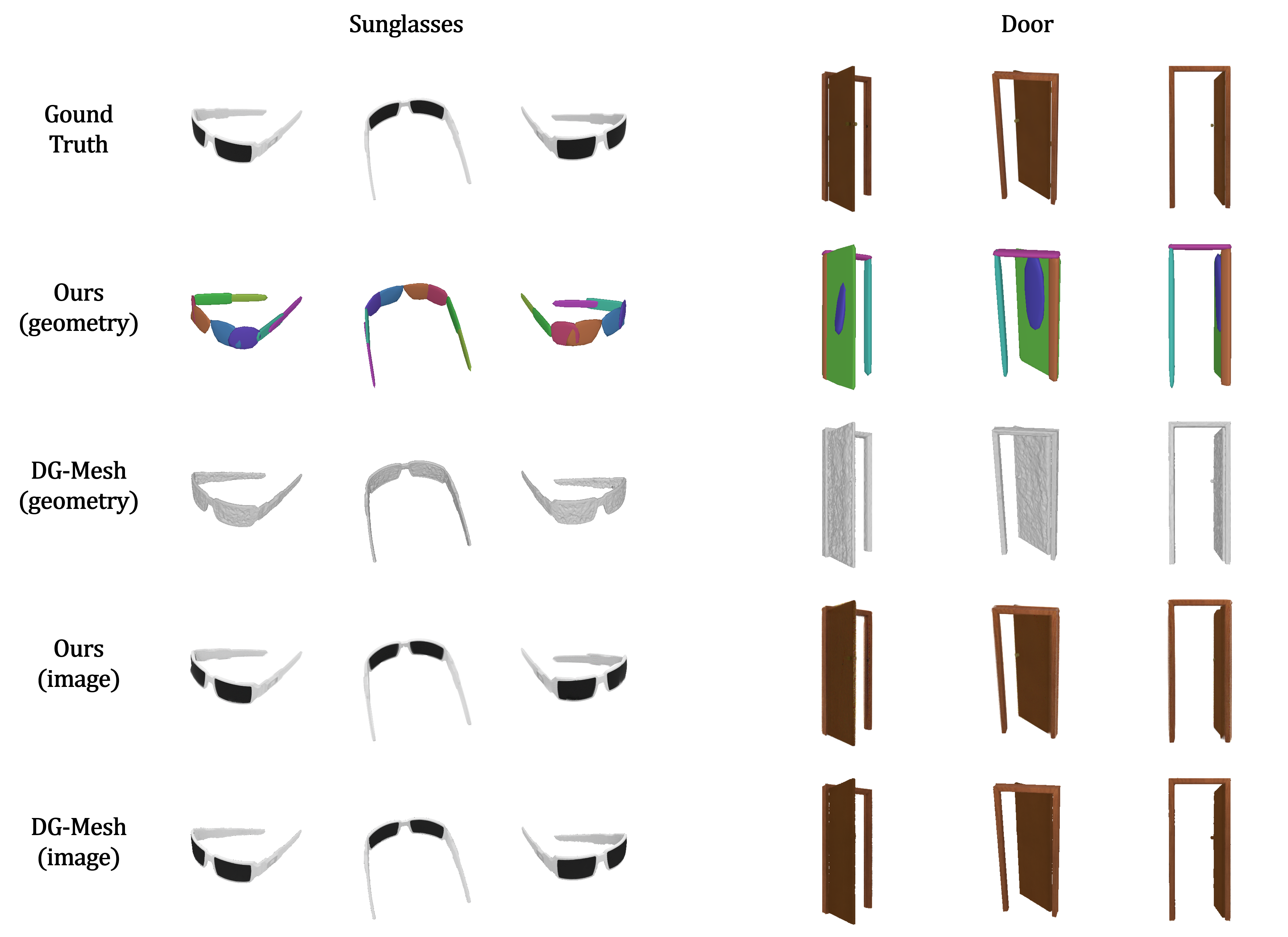}
   \caption{\textbf{Visualization Results for Dynamic Primitive Dataset.} We show both geometry results and rendering results.}
   \label{fig:pri_data_vis3}
\end{figure*}

We present visual results for all cases from Dynamic Primitive Dataset in Fig. \ref{fig:pri_data_vis1}, \ref{fig:pri_data_vis2}, \ref{fig:pri_data_vis3}. Additionally, we provide detailed calculation procedures for the evaluation metrics in the following subsections. We also provide the PSNR and SSIM rendering metrics for the geometry-based methods tested on Dynamic Primitive Dataset, as detailed in Tab. \ref{tab:motion_geo_rendering}.

\subsection{Details for Structured Motion Tracking Accuracy}
To compute structured motion tracking accuracy, we first select a canonical ground truth mesh at one timestamp. We sample 50,000 points on it using the \textit{trimesh} sampling function (with a fixed seed) and record their barycentric coordinates. We then use these coordinates and the ground truth meshes from all other timestamps to compute the points' corresponding positions over time. This process yields the motion tracking ground truth results.

Next, we use the methods under evaluation to predict the positions of these 50,000 points over time. For our method, we first obtain our primitive geometry at the canonical timestamp corresponding to the initial sampling, bind each point to its nearest primitive, and then calculate the point's motion from that primitive's motion parameters. For methods that provide consistent meshes, we use a similar approach by binding each point to its nearest triangular face and computing its motion from the face's motion. For methods that do not provide consistent mesh results but provide deformation networks, we first use their inverse network to map the points to their respective canonical space, and then use their forward network to map these points to each observation space. For Shape of Motion, we compute the point motion using their officially provided procedure.

\subsection{Details for Structured Dynamic Reconstruction Results}
For traditional mesh quality metrics, we impose a special dynamic constraint. Evaluated methods must provide consistent meshes, where the geometry at each frame shares an identical vertex count and face definition. This means the vertex indices for any given face must remain fixed over time. After obtaining these per-frame consistent results, we compute the traditional metrics against the per-frame ground truth mesh. The final dynamic reconstruction results are the average of these metrics across all timestamps.

\section{Visualization Results for D-NeRF Dataset}
\begin{figure*}[!t]
  \centering
   \includegraphics[width=1\linewidth]{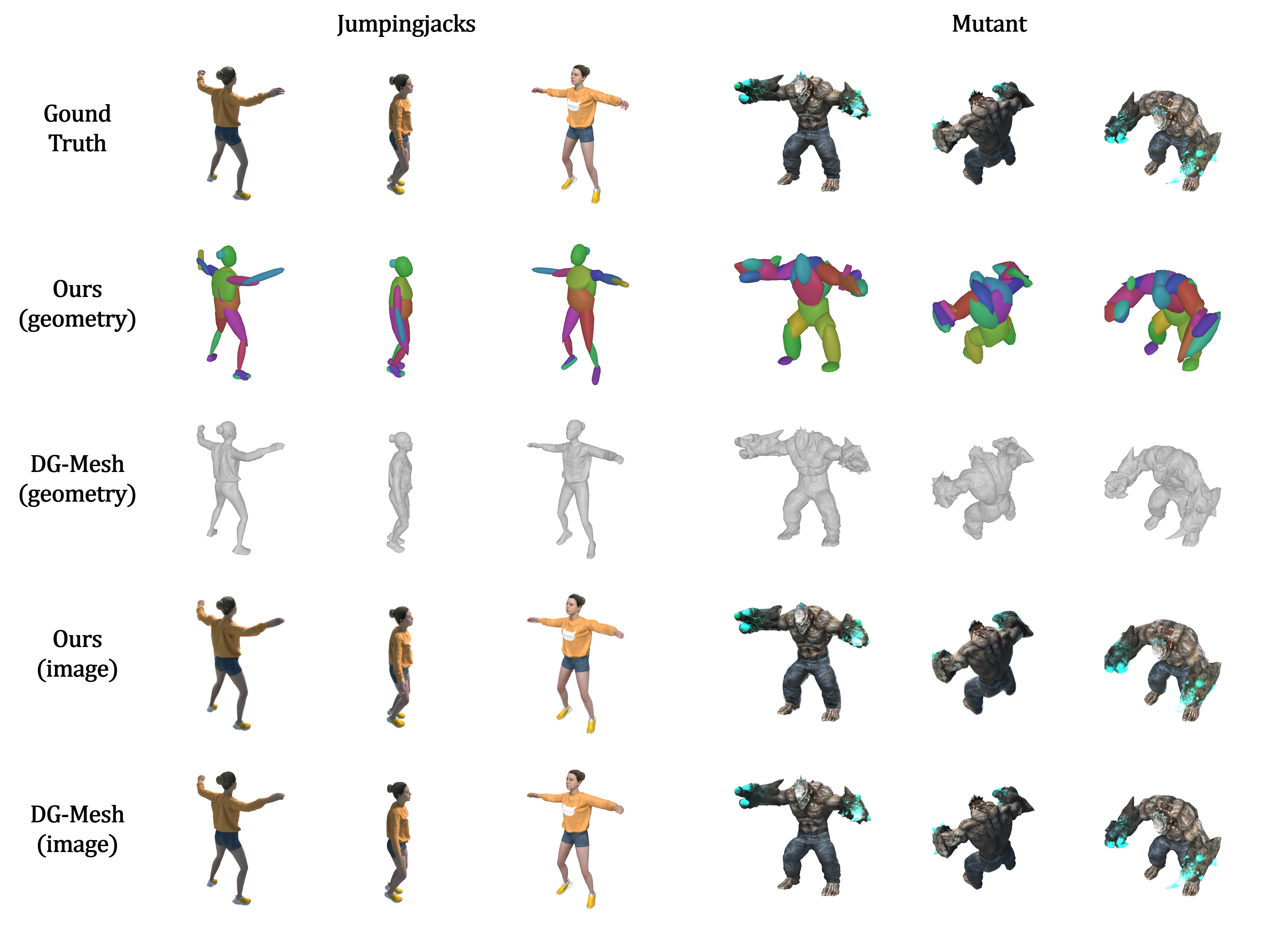}
   \caption{\textbf{Visualization Results for D-NeRF Dataset.} We show both geometry results and rendering results.}
   \label{fig:dnerf_data_vis1}
\end{figure*}

\begin{figure*}[!t]
  \centering
   \includegraphics[width=1\linewidth]{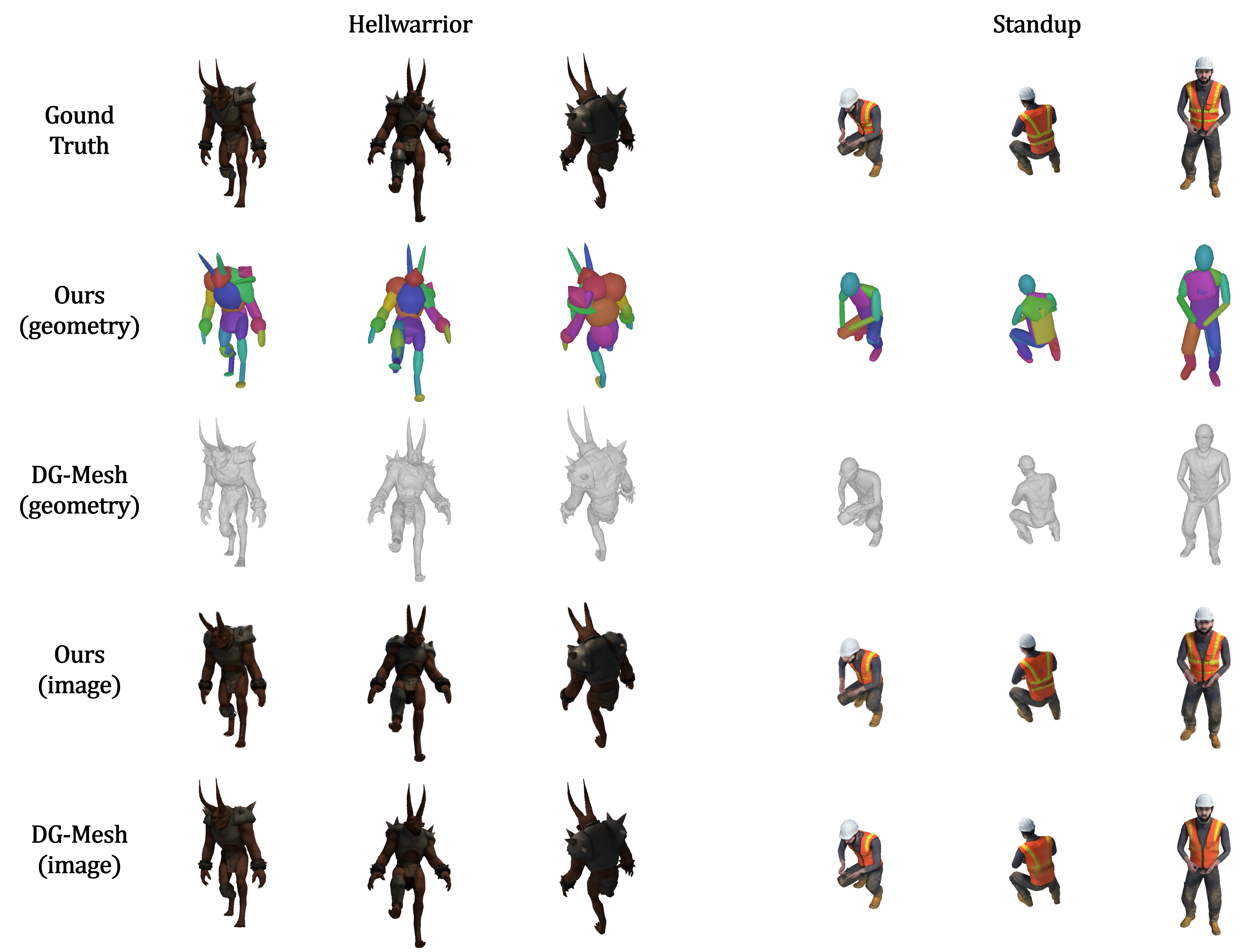}
   \caption{\textbf{Visualization Results for D-NeRF Dataset.} We show both geometry results and rendering results.}
   \label{fig:dnerf_data_vis2}
\end{figure*}

\begin{figure}[!t]
  \centering
   \includegraphics[width=1\linewidth]{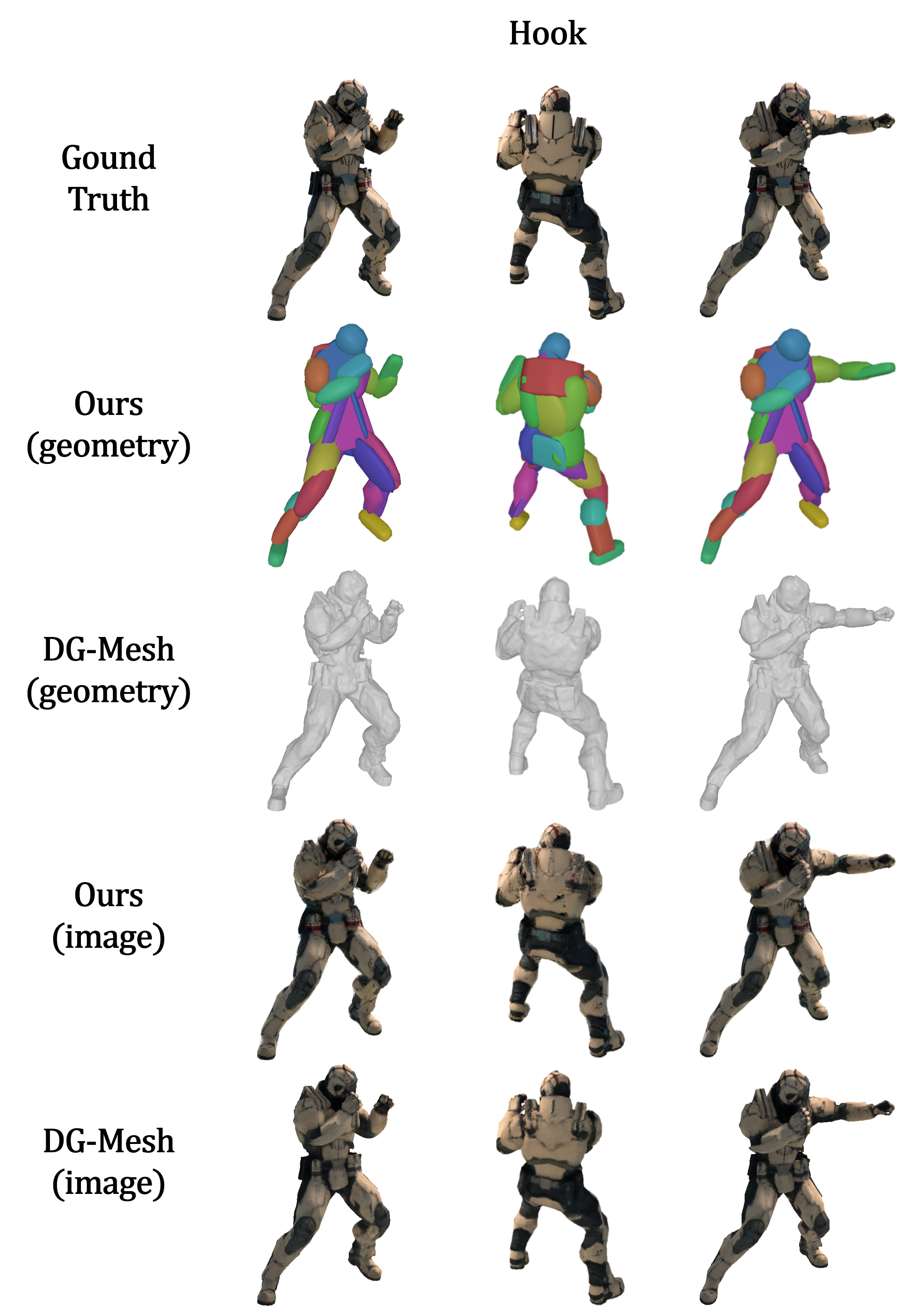}
   \caption{\textbf{Visualization Results for D-NeRF Dataset.} We show both geometry results and rendering results.}
   \label{fig:dnerf_data_vis3}
\end{figure}

Here we present the complete visualization results for sequences we have used in D-NeRF Dataset, including structured dynamic geometry results and rendering results, as detailed in Fig. \ref{fig:dnerf_data_vis1}, \ref{fig:dnerf_data_vis2}, \ref{fig:dnerf_data_vis3}. The structured results make the object more interpretable and facilitate further operations on the object's parts, such as articulation and editing.

\section{Visualization Results for Real World Data}
\begin{figure}[!t]
  \centering
   \includegraphics[width=1\linewidth]{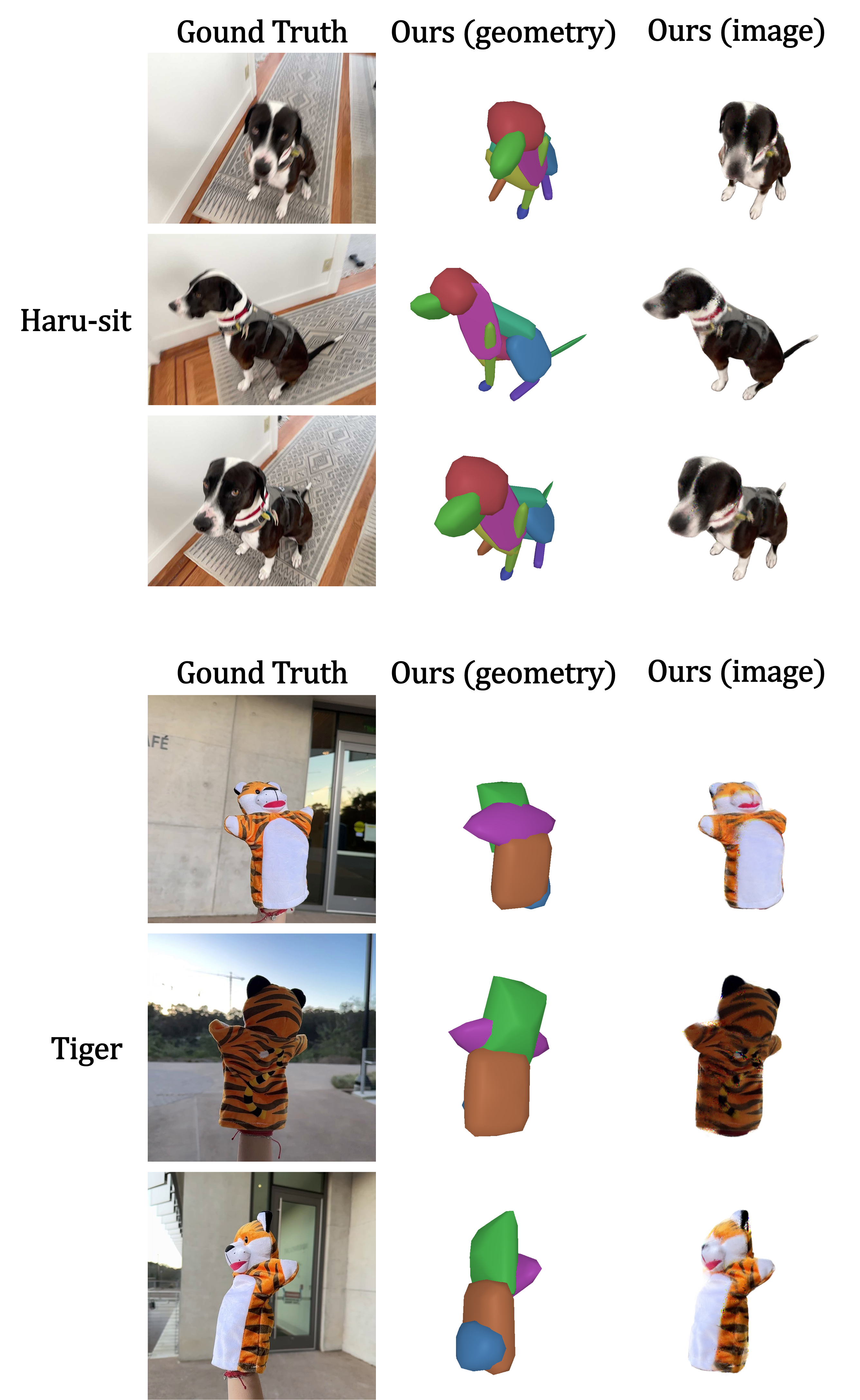}
   \caption{\textbf{Visualization Results for Real World Data.} We show both geometry results and rendering results.}
   \label{fig:real_data_vis}
\end{figure}

\begin{figure*}[!t]
  \centering
   \includegraphics[width=1\linewidth]{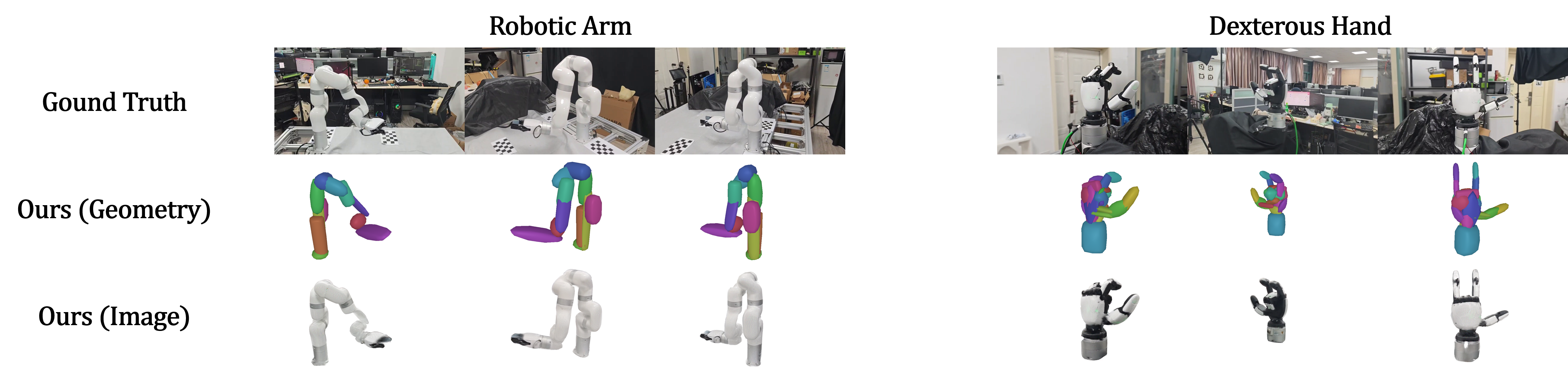}
   \caption{\textbf{Visualization Results for Real World Robotic Scenarios.} We show both geometry results and rendering results.}
   \label{fig:real_rob_data_vis}
\end{figure*}

We also test our method on real-world scenarios. Here we select and test two cases from Self-Captured iPhone Dataset \cite{liu2024dynamic_supp} and DyCheck Dataset \cite{NEURIPS2022_dab5a29f_supp}, visualizing their results in Fig. \ref{fig:real_data_vis}. These causal video cases, which feature sparse viewpoints and imprecise annotations, are indeed challenging for our method. Additionally, we have recorded two dynamic video sequences of real world robotic scenarios, including a moving robotic arm and a dexterous hand. These videos were processed following the data processing pipeline from \cite{liu2024dynamic_supp}. With adequate viewpoint coverage, our method performs well on these sequences, as shown in Fig. \ref{fig:real_rob_data_vis}.

\section{Visualization for Applications}
\begin{figure*}[t]
    \centering
    
    \begin{subfigure}{\linewidth} 
        \centering
        \includegraphics[width=\linewidth]{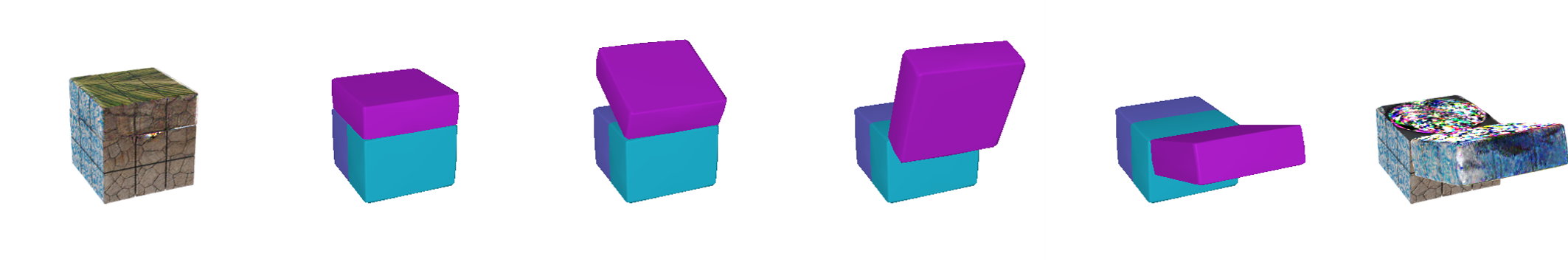}
        \caption{From Treasure Box to Rubik's Cube.}
    \end{subfigure}
    \begin{subfigure}{\linewidth}
        \centering
        \includegraphics[width=\linewidth]{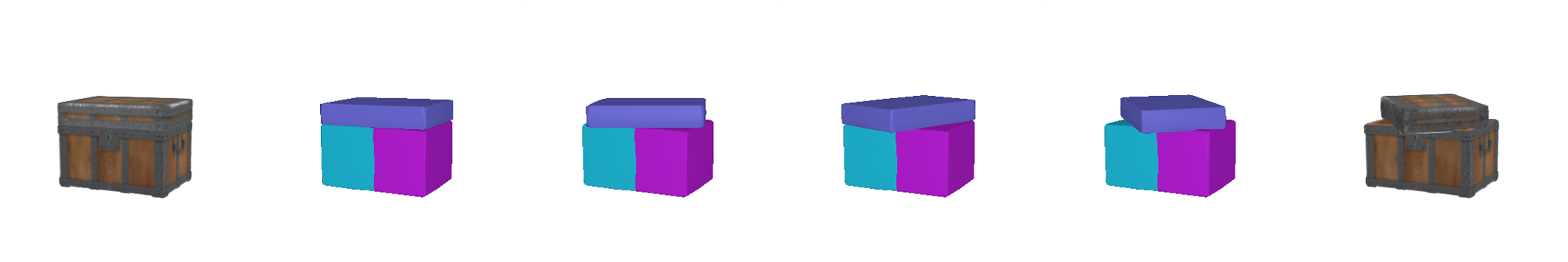}
        \caption{From Rubik's Cube to Treasure Box.}
    \end{subfigure}
    \caption{\textbf{Visualization for Motion Pattern Transfer.} Here we swap the original motions of the Treasure Box and the Rubik's Cube, assigning new motion patterns to both dynamic objects.}
    \label{fig:motion_switch}
\end{figure*}

\begin{figure*}[!t]
  \centering
   \includegraphics[width=1\linewidth]{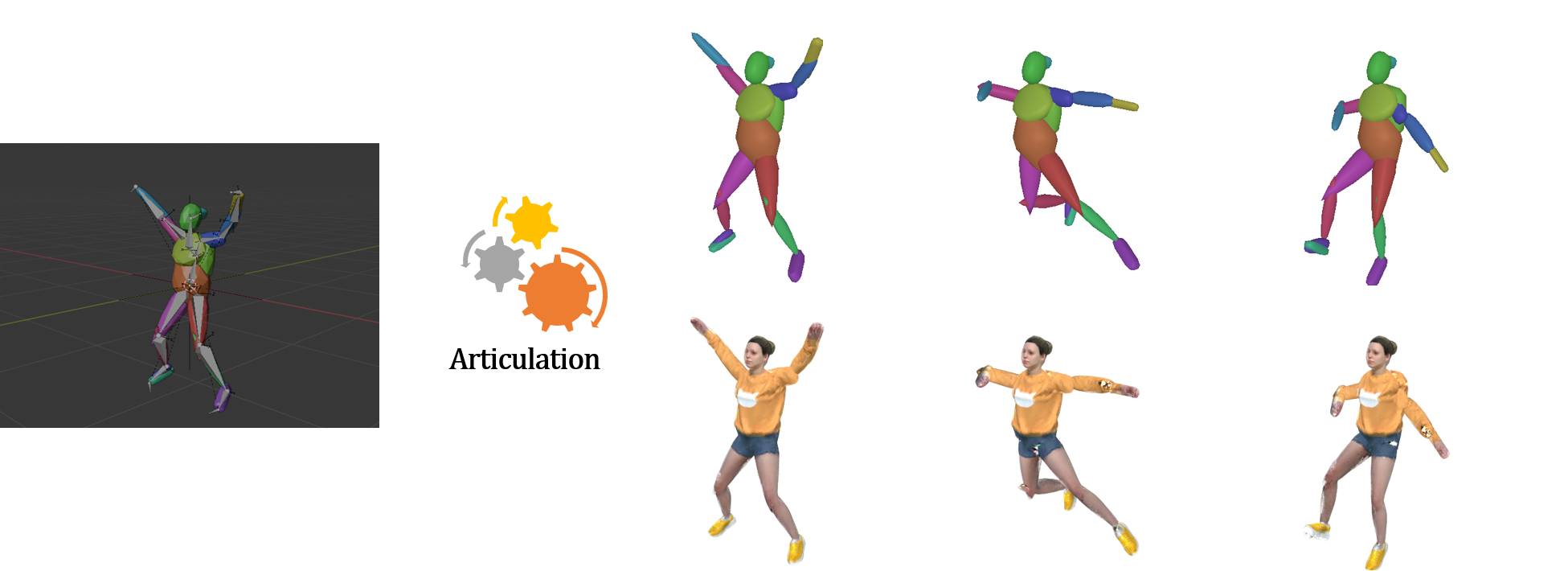}
   \caption{\textbf{Visualization of Easy Articulation.} After simple operations, such as simple skeleton annotation in software like Blender, we can obtain a structured articulable representation, enabling free articulation editing.}
   \label{fig:human_articulate}
\end{figure*}

Our method provides a structured geometric representation that enables easy editing of the reconstructed results. First, as demonstrated in the teaser, we can easily perform motion pattern transfer. For instance, we can transfer the opening motion of the Treasure Box lid to the top layer of the Rubik's Cube, or apply the rotation of the Rubik's Cube to the lid, as shown in Fig. \ref{fig:motion_switch}.

We can also perform various physics-based motion simulations on individual object parts. For instance, we simulate the Rubik's Cube rotating upon impact and stopping due to friction, or the Treasure Box lid accelerating open under force and slowly settling back due to joint damping. Visualizations of this application are available in our supplementary video.

Additionally, we can articulate our humanoid reconstructions. Leveraging our structured representation, we can control limb movements in Blender by simply setting up a skeleton and Inverse Kinematics (IK) parameters based on the structured representation, eliminating the need for skinning, as shown in Fig. \ref{fig:human_articulate}. 

Comprehensive visualizations of all these applications are provided in our supplementary video.




\end{document}